\documentclass[letterpaper]{article} % DO NOT CHANGE THIS
\usepackage[preprint]{aaai2027}  % DO NOT CHANGE THIS
\usepackage[hyphens]{url}  % DO NOT CHANGE THIS
\usepackage{graphicx} % DO NOT CHANGE THIS
\usepackage{natbib}  % DO NOT CHANGE THIS AND DO NOT ADD ANY OPTIONS TO IT
\usepackage{caption} % DO NOT CHANGE THIS AND DO NOT ADD ANY OPTIONS TO IT
\usepackage{algorithm}
\usepackage{algorithmic}
\usepackage{amsmath}
\usepackage{amssymb}
\usepackage{booktabs}
\usepackage{array}

\usepackage[table]{xcolor} % loads colortbl for \rowcolor; keep before tikz to avoid an option clash
\usepackage{tikz}
\definecolor{accblue}{HTML}{0F4D92}
\definecolor{accred}{HTML}{B64342}
\definecolor{secrowgray}{gray}{0.97} % very light gray for table section headers
\newcommand{\glyphacc}{\tikz[baseline=-0.5ex]{\draw[accblue,line width=1.5pt](0,0)--(1.5em,0);\node[circle,fill=accblue,draw=white,line width=0.5pt,minimum size=6pt,inner sep=0]at(0.75em,0){};}}
\newcommand{\glyphnll}{\tikz[baseline=-0.5ex]{\draw[accred,line width=1.5pt,dash pattern=on 3pt off 2pt](0,0)--(1.5em,0);\node[fill=accred,draw=white,line width=0.5pt,minimum size=5.5pt,inner sep=0]at(0.75em,0){};}}

\makeatletter
\patchcmd{\maketitle}
  {\insert\aaai@copyrightins}
  {\insert\footins}
  {}
  {\PackageError{RecurTrace}{Could not install the AAAI submission-notice layout fix}
    {Check for an updated aaai2027.sty before compiling.}}
\makeatother

\title{RecurTrace: Adaptive Latent Reasoning with Loop-Time Memory}

\author{
    Yuxiang Wang\textsuperscript{\mdseries 1}\quad
    Kunyu Feng\textsuperscript{\mdseries 1}\quad
    Yingda Shen\textsuperscript{\mdseries 1}\quad
    Haoning Xu\textsuperscript{\mdseries 4}\quad
    Junyu Wang\textsuperscript{\mdseries 5}\quad
    Zhizheng Wu\textsuperscript{\mdseries 1,2,3}\quad
}
\affiliations{
    \textsuperscript{1}The Chinese University of Hong Kong, Shenzhen\\
    \textsuperscript{2}Shenzhen Loop Area Institute\quad
    \textsuperscript{3}Amphion Technology Co., Ltd.\\
    \textsuperscript{4}The Chinese University of Hong Kong\quad
    \textsuperscript{5}Tianjin University\\
    \texttt{yuxiangwang1@link.cuhk.edu.cn}
}

\begin{document}

\maketitle

\begin{abstract}
    Repeating a small block of middle layers increases a language model's effective inference depth without adding parameters or generating extra tokens, and recent work shows that this latent recurrence improves reasoning. However, two design choices limit these gains. Each iteration sees only the previous output and cannot directly access earlier computations. Moreover, a fixed loop count wastes depth on easy inputs while leaving hard ones with too little computation. We introduce \textbf{RecurTrace}\footnote{Code: \url{https://github.com/JensenYX/RecurTrace}}, which addresses both limitations using the loop's own trajectory. Specifically, Loop Memory Attention lets each looped layer attend to its own states from previous iterations along the loop-time axis, so the model can revisit earlier computations instead of relying on the latest state alone. A halting head then reads the loop state and predicts whether to continue, with supervision from an oracle that identifies when additional depth still reduces loss. In a controlled MathQA comparison on the same looped backbone, RecurTrace achieves $56.9\%$ accuracy with an average of $2.0$ loops, exceeding the best fixed loop depth by $2.2$ points at matched compute. By comparison, ACT and PonderNet collapse to one loop, and CALM reaches only $54.1\%$ with $5.6$ loops, while the stronger LoopUS-Conf and TaH-Mismatch baselines reach $55.3\%$ at $3.2$ loops and $55.7\%$ at $2.1$ loops. Finally, RecurTrace improves generation accuracy over same-budget fine-tuned baselines at $0.6$B, $1.7$B, $4$B, and $8$B, with the gain growing with model size from $0.6$ to $3.4$ points.
    \end{abstract}

\begin{figure}[t]
\centering
\includegraphics[width=0.99\columnwidth]{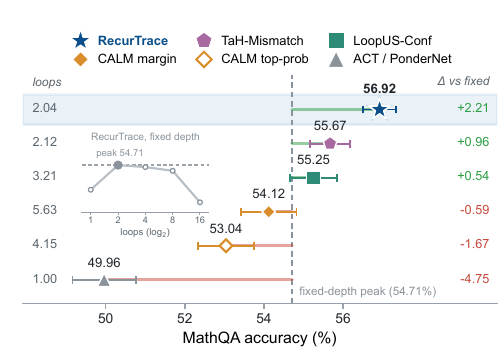}
\caption{MathQA accuracy versus mean loop count by halting method ($1.7$B, $3$ training $\times$ $8$ evaluation seeds). Points are means $\pm 1$ SD across training seeds; the dashed line marks the best fixed depth ($54.71\%$). RecurTrace reaches $56.92\%$ at ${\sim}2$ loops, $1.25$ points above TaH-Mismatch. Inset: fixed-depth accuracy peaks at $2$ loops.}
\label{fig:pareto}
\end{figure}

\section{Introduction}
A transformer spends the same number of layers on an easy question and on a hard one. This is convenient for training but wasteful at inference. Reasoning quality scales with effective depth, whether the depth comes from stacking more layers, from spelling out a chain of thought in tokens \citep{wei2022cot}, or from iterating a block of layers in latent space \citep{saunshi2025looped,geiping2025recurrent}. Among these routes, latent looping increases effective depth without emitting additional reasoning tokens. It adds depth at inference without new parameters or longer outputs, and it leaves the tokenizer and the decoding loop unchanged.

Two problems keep latent looping from delivering on this promise. The first is that the loop forgets. A looped block re-runs the same layers on their own output and passes forward only a single hidden state. Whatever an earlier iteration computed must fit into that state or be overwritten by later passes. A loop that cannot consult its own past cannot reason across iterations. The second problem is that the loop cannot budget. Looped models fix the iteration count for the whole dataset, so an easy question gets as many loops as a hard one. Figure~\ref{fig:pareto} shows the cost. On MathQA, a $1.7$B looped model peaks at two loops with $54.7\%$ accuracy, stays near $53.6\%$ at the eight-loop training clamp, and falls to $47.5\%$ at sixteen. More loops soon stop helping and then hurt, so no fixed budget suits every input.

For the memory problem, two concurrent efforts attach external stores to looped language models, gated key-value banks in \citet{frey2026adaptive} and a multi-slot buffer with read-write routers in MeSH \citep{yu2025mesh}, each adding extra parameters and a learned read-write mechanism. For budgeting, Adaptive Computation Time (ACT) \citep{graves2016act} and PonderNet \citep{banino2021pondernet} halt under a compute penalty or a geometric prior, confidence rules such as CALM \citep{schuster2022calm} exit once an intermediate state looks certain, and Mixture-of-Recursions gives each token its own recursion depth but trains its routers from scratch \citep{bae2025mor}. On a pretrained model whose block is looped, a penalty on extra loops makes halting stop too early, and a confidence threshold makes it stop too late.

Our starting point is that the loop already produces everything both problems require, so no outside machinery is needed. The states it computed on earlier passes are a ready-made memory, and the state it holds now carries evidence about whether another pass will change the answer. Therefore, we propose \textbf{RecurTrace}, which converts a pretrained Qwen3 model \citep{yang2025qwen3} into a memory-augmented looped reasoner with learned halting. For the forgetting problem, we give every looped layer a Loop Memory Attention module. Along a loop-time axis, each token attends from its current state to its own states from previous loops, so a layer can read what it computed one or two iterations earlier. The module sits beside the existing self-attention, adds few parameters, and mixes no information across token positions, so it cannot leak future tokens. For the budgeting problem, we train a halting head that reads the loop state and predicts whether a deeper loop will still help. Its supervision comes from an oracle that records, for each training example, the depth at which the answer stops improving, a measured target used directly, with no added compute penalty or confidence proxy. At inference the model loops until the head says stop, so depth is set per input rather than fixed across the dataset.

The memory and the halting head need each other. Without the memory, extra loops barely help, so the halting head gains little by running an input longer. Without the halting head, every input gets the same number of loops, so the extra loops the memory makes useful are wasted on easy inputs and denied to hard ones. Figure~\ref{fig:pareto} previews the outcome on MathQA. RecurTrace reaches $56.9\%$ at $2.0$ loops, improving on the best fixed loop budget by $2.2$ points at a comparable mean depth and beating every adaptive-compute baseline we test, including the recent LoopUS \citep{park2026loopus} and Think-at-Hard \citep{fu2026tah}.

RecurTrace is also practical to adopt. All base weights stay frozen, the added modules hold at most about $2.2\%$ of the parameters, and they start out nearly inactive, so a single loop reproduces the pretrained model exactly, a safe floor that more loops build on. Across $0.6$B, $1.7$B, $4$B, and $8$B, RecurTrace raises generation accuracy over same-budget baselines at every scale, with the gain growing to $3.4$ points at $8$B, and lowers teacher-forced loss at all four. Our contributions are:
\begin{itemize}
\item \textbf{Loop Memory Attention.} An attention along the loop-time axis through which every token reads its own states from earlier loops. The looped block stays fully weight-tied and carries no external memory store.
\item \textbf{Oracle-distilled halting.} A sequence-level halting head supervised by whether a deeper loop still lowers the loss, not by a prior, penalty, or confidence rule. It matches or exceeds the best fixed depth across six tasks.
\item \textbf{One model for every depth.} One trained model serves every 
test-time loop depth, not one 
model for each budget, and the recipe holds from 
$0.6$B to $8$B with accuracy and likelihood 
gains over same-budget baselines.
\end{itemize}

\section{Related Work}
\paragraph{Looped and recurrent-depth transformers.}
Reusing layers to add depth without new parameters dates to the Universal Transformer \citep{dehghani2019universal}, weight-tied models such as ALBERT \citep{lan2020albert}, and deep equilibrium models \citep{bai2019deq}. It pays off for reasoning, where \citet{saunshi2025looped} show a looped $k$-layer block can rival a $kL$-layer network, \citet{geiping2025recurrent} sample the iteration count to reach deeper unrolls, and \citet{zhu2025ouro} pretrain looped models at scale. Encode-Think-Decode \citep{koishekenov2026etd} re-runs a pretrained middle block, and CoTFormer \citep{mohtashami2025cotformer} interleaves looped and fresh representations. Plain-looping approaches pass only the previous output between iterations and keep no addressable record of intermediate states. RecurTrace adds that record and reads it with attention along the loop-time axis. Dreamer \citep{knupp2026dreamer} also attends along depth in a recurrent sparse-expert architecture, whereas RecurTrace grafts its loop-time read and halting onto a frozen dense model.
% Because the memoryless re-iteration of ETD matches our plain-looping ablation, Table~\ref{tab:memory} doubles as a direct comparison with this closest prior method.

External stores sit outside the recurrence, including the gated memory banks of \citet{frey2026adaptive} and the multi-slot buffer of MeSH \citep{yu2025mesh}. DiscoLoop \citep{fu2026discoloop} and PonderLM \citep{zeng2026ponderlm} instead feed re-encoded predictions back into the loop. RecurTrace keeps the block weight-tied and adds no store, querying only the trajectory the recurrence already produced. The concurrent LoopUS \citep{park2026loopus} likewise recasts a pretrained model into encode-think-decode blocks with a selective gate, keeps no addressable history, and exits on a confidence rule we find overspends the loop budget.

\begin{figure*}[t]
    \centering
    \includegraphics[width=\textwidth]{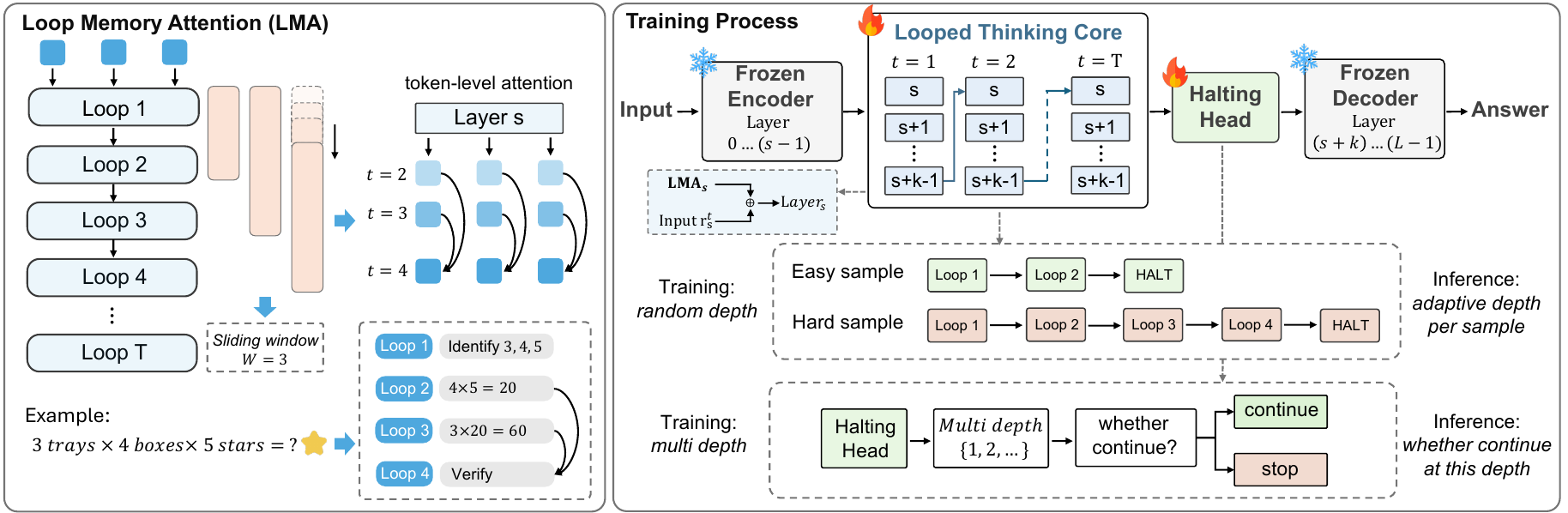}
    \caption{RecurTrace architecture. A frozen encoder feeds a weight-tied block that recurs to variable depth, re-injecting its input after the first loop. Before each looped layer, Loop Memory Attention reads same-position states from a three-loop sliding window. Random-depth training supports multiple depths, while an oracle-distilled halting head predicts whether another loop will help, allocating more computation to hard inputs. After halting, the remaining frozen layers generate the answer.}
    \label{fig:arch}
    \end{figure*}

\paragraph{Adaptive computation and early exit.}
Adaptive Computation Time \citep{graves2016act} accumulates a stopping probability, and PonderNet \citep{banino2021pondernet} halts under a geometric prior over steps. For pretrained models, depth-adaptive decoding \citep{elbayad2020depth}, DeeBERT \citep{xin2020deebert}, and CALM \citep{schuster2022calm} exit early once a layer is confident, while Mixture-of-Depths \citep{raposo2024mod} spends each layer on only some tokens. Mixture-of-Recursions adapts the recursion depth per token with scratch-trained routers \citep{bae2025mor}, and Think-at-Hard \citep{fu2026tah} iterates only tokens predicted wrong after one pass. We instead adapt how often the looped block repeats, halt per sequence, and supervise the head with an oracle marking whether a deeper loop lowers the loss, not with a prior, penalty, confidence, or token-mismatch rule, so it asks when added depth pays off for the sequence rather than whether one token was wrong.

\paragraph{Scaling test-time compute.}
Chain-of-thought prompting \citep{wei2022cot}, self-consistency \citep{wang2023selfconsistency}, and adaptive test-time scaling \citep{zhai2026adaptive} use extra decoding compute; reinforcement learning trains reasoning models \citep{zhang2025rlreasoning}. A complementary line stays latent, appending pause tokens \citep{goyal2024pause} or feeding hidden states back as continuous thoughts \citep{hao2025coconut,wei2025simcot}. Latent looping scales per-token computation without emitting tokens, and RecurTrace asks how to make each loop count and when to stop.

\section{Method}
RecurTrace starts from a pretrained LLM with $L$ transformer layers, where layer $\ell$ maps a hidden state $h$ to $\mathrm{layer}_\ell(h)$. We designate a contiguous block of $k$ middle layers, $\mathcal{B}=\{s,\dots,s{+}k{-}1\}$, as a looped thinking block and repeat it $T$ times at inference. The layers before the block act as an encoder and the layers after it as a decoder, following the encode-think-decode view \citep{koishekenov2026etd}. Figure~\ref{fig:arch} shows the design. Three ingredients make the loop useful. A memory carries information across iterations, variable-depth training exposes the model to many depths, and a halting head picks the depth per input.

\paragraph{Choosing the looped block.}
We place the block where iterating helps most. Following the angular-distance probe of \citet{koishekenov2026etd}, we first locate the broad middle span whose representations move least between adjacent layers, the span that behaves most like a fixed point, and then pick a compact three-layer window inside it by behavioral ablation. This yields layers $12$--$14$ for the $28$-layer $0.6$B and $1.7$B Qwen3 bases and layers $15$--$17$ for the $36$-layer $4$B and $8$B ones. Probe-core and late-block controls are comparable or weaker (Appendix~\ref{app:block}).

\paragraph{Looped computation.}
Let $e$ be the block input, the output of layer $s{-}1$. We write $x_\ell^{(t)}$ for the state produced by layer $\ell$ on loop $t$. Each loop after the first begins by re-injecting the block input,
\begin{equation}
h \leftarrow h + \alpha \, \mathrm{RMSNorm}(e),
\label{eq:inject}
\end{equation}
with a scalar $\alpha$ initialized to zero in the ReZero style \citep{bachlechner2021rezero}, so injection starts as a no-op. Re-injecting the input anchors the recurrence as depth grows \citep{geiping2025recurrent}, a device with roots in the deep-thinking and recall networks of \citet{schwarzschild2021deepthinking} and \citet{bansal2022recall}. Inside the loop, before applying each block layer we add a memory term and then run the unchanged layer,
\begin{align}
h &\leftarrow h + \mathrm{LMA}_\ell\!\left(x_\ell^{(t-1)}, M_\ell^{(t)}\right),
\label{eq:lma}\\
x_\ell^{(t)} &\leftarrow \mathrm{layer}_\ell(h),
\end{align}
where $M_\ell^{(t)}=\{x_\ell^{(t')} \mid \max(1,t{-}W)\le t' \le t{-}1\}$ is a sliding window of the layer's own outputs from up to $W$ earlier loops. On the first loop $M_\ell$ is empty, so $\mathrm{LMA}_\ell$ returns zero, and at initialization $T{=}1$ reproduces the base model.

\begin{algorithm}[t]
    \small
    \caption{Difficulty-adaptive looped inference}
    \label{alg:halt}
    \textbf{Input} tokens, with budget $T_{\max}$, floor $T_{\min}$, and threshold $\tau$\\
    \textbf{Output} next-token distribution
    \begin{algorithmic}[1]
    \STATE $h \leftarrow$ encode the input through layers $0..s{-}1$, then set $e \leftarrow h$
    \FOR{$t = 1$ \TO $T_{\max}$}
    \STATE \textbf{if} $t > 1$ \textbf{then} $h \leftarrow h + \alpha\,\mathrm{RMSNorm}(e)$ \hfill(input injection)
    \FOR{$\ell \in \mathcal{B}$}
    \STATE \textbf{if} $t > 1$ \textbf{then} $h \leftarrow h + \mathrm{LMA}_\ell(x_\ell^{(t-1)}, M_\ell^{(t)})$
    \STATE $h \leftarrow \mathrm{layer}_\ell(h)$, then store $x_\ell^{(t)} \leftarrow h$ in $M_\ell$
    \ENDFOR
    \STATE $p_t \leftarrow \sigma(w^{\!\top}\mathrm{pool}(h)+b)$ \hfill(continue probability)
    \STATE \textbf{if} $t \ge T_{\min}$ \textbf{and} $p_t < \tau$ \textbf{then break} \hfill(easy input, stop early)
    \ENDFOR
    \STATE \textbf{return} decode $h$ through layers $s{+}k..L{-}1$ and the LM head
    \end{algorithmic}
    \end{algorithm}

\paragraph{Loop Memory Attention.}
The memory term in Eq.~\eqref{eq:lma} is an attention over the loop-time axis. For a query state $q=x_\ell^{(t-1)}$ and a window of $m$ past states stacked as $X$, we normalize both, form per-head queries, keys, and values $Q,K,V$ with QK-normalization, and for every token position attend across the $m$ loop slots,
\begin{equation}
A_{i,j} = \mathrm{softmax}_j\!\left(\frac{Q_i^{\!\top} K_{i,j}}{\sqrt{d_h}} - \beta_h\,(t - t_j)\right),
\label{eq:attn}
\end{equation}
\begin{equation}
\mathrm{LMA}_\ell(q,X) = \gamma \cdot g \odot \Big( \big[\textstyle\sum_j A_{i,j} V_{i,j}\big]_i W_o \Big).
\label{eq:gate}
\end{equation}
The term $-\beta_h (t-t_j)$ is a relative loop-distance bias with a free, signed, learnable per-head slope $\beta_h$, initialized to ALiBi values \citep{press2022alibi}. Because the bias depends only on how many loops ago a state was written, the module extrapolates to more loops than it saw in training. The attention runs along loop time only and never mixes token positions, so it adds no path for future tokens to leak into the present. A scalar gate $\gamma$ and a token-wise gate $g=\sigma(\mathrm{MLP}([q,\overline{X}]))$, where $\overline{X}$ is the mean of the memory, let the model close the memory on inputs that do not need it and open it on inputs that do. The combined gate starts near zero, so the memory path opens gradually during training. Reading the past through attention, rather than through an external buffer, also keeps the additions close to the pretrained model. The block stays weight-tied, no slot routing is learned, and the sliding window caps the cost of remembering at a constant.

\paragraph{Variable-depth training.}
We do not fix the loop count during training. On each forward pass we draw $T$ from a clamped log-normal Poisson distribution \citep{geiping2025recurrent}, which places most mass at small depths but keeps a heavy tail of deep unrolls, and we apply the language-model loss at the sampled depth. The realized distribution has a mean depth near $3.9$, and about a fifth of passes run six or more loops (Appendix~\ref{app:vardepth}). This single change lets one trained model serve any test-time depth and extrapolate beyond the depths it trained on. All loop parameters are added to the pretrained weights, and the nearly closed gates keep the model close to its starting point early in training.

\paragraph{Difficulty-adaptive halting.}
To decide how deep to go, we add a lightweight halting head. After loop $t$ we pool the block state over prompt positions into a vector $z_t$ and predict a continue probability $p_t=\sigma(w^{\!\top} z_t + b)$. We supervise it by distilling an oracle that reads the same forward pass. Let $\mathcal{L}_t$ be the per-example loss from decoding at depth $t$. The oracle marks ``continue'' only when a deeper loop helps,
\begin{equation}
y_t = \mathbb{1}\!\left[\min_{t' > t} \mathcal{L}_{t'} < \mathcal{L}_t - \delta \right],
\label{eq:oracle}
\end{equation}
for a margin $\delta$, and the head is trained with binary cross-entropy against $y_t$ at a set of probe depths. The same head also supports the ACT and PonderNet objectives, which we use for the baselines. At inference (Algorithm~\ref{alg:halt}) the model loops until the head says stop or a budget $T_{\max}$ is reached, with a floor of $T_{\min}$ loops. The stopping threshold $\tau$ is chosen on held-out data, taking the smallest mean depth that does not lose likelihood, and is never tuned on the test set.

\begin{table}[t]
    \centering
    \small
    \setlength{\tabcolsep}{4pt}
    \begin{tabular*}{\columnwidth}{@{\extracolsep{\fill}}l c c c l@{}}
    \toprule
    \textbf{Scale} & \textbf{\boldmath$L$} & \textbf{\boldmath$d_{\text{model}}$} & \textbf{Loop block} & \textbf{Training} \\
    \midrule
    $0.6$B & 28 & 1024 & layers $12$--$14$ & loop modules \\
    $1.7$B & 28 & 2048 & layers $12$--$14$ & loop modules \\
    $4$B & 36 & 2560 & layers $15$--$17$ & loop modules \\
    $8$B & 36 & 4096 & layers $15$--$17$ & loop modules \\
    \bottomrule
    \end{tabular*}
    \caption{Per-scale models. $L$ is the number of base layers and $d_{\text{model}}$ the hidden size. Every scale freezes the base and trains only the loop, memory, injection, and halting parameters, with per-scale training schedules in Appendix~\ref{app:impl}.}
    \label{tab:app-models}
    \end{table}

\section{Experiments}
Our experiments answer three questions about whether learned halting beats the best fixed depth, whether the loop-memory gains carry across model scale and a broad task set, and what the extra loops cost.

\subsection{Experimental Setup}
\paragraph{Setup.}
We build RecurTrace on four Qwen3 scales (Table~\ref{tab:app-models}). At every scale we add the looped block to a pretrained checkpoint, freeze the base weights, and train only the loop, memory, injection, and halting parameters. Every comparison is against a no-loop baseline under a matched budget of steps, tokens, data, and hardware. Because the baseline fine-tunes the full model while RecurTrace adds at most about $2.2\%$ trainable parameters, any budget mismatch favors the baseline. The training mixture holds about $1.15$ million direct-answer examples combining synthetic generators with real math, multi-hop question answering, and controlled logic (Appendix~\ref{app:impl}). Since every target is a short final answer, test-time loops add latent computation rather than output tokens. We evaluate a \textbf{train-covered} reasoning suite of $14$ tasks including GSM8K \citep{cobbe2021gsm8k}, MATH \citep{hendrycks2021math}, and MathQA \citep{amini2019mathqa}, and a \textbf{classic} suite of eight general benchmarks including ARC \citep{clark2018arc}, HellaSwag \citep{zellers2019hellaswag}, and MMLU \citep{hendrycks2021mmlu}, reporting greedy-decoding generation accuracy and teacher-forced NLL (nats, lower is better). BIG-Bench Hard \citep{suzgun2023bbh} is diagnostic only.

We run the controlled adaptive-halting comparison on MathQA at $1.7$B, the setting with the most room for adaptive depth. Direct-answer GSM8K scores low at this scale and most classic suites peak at shallow depth, so both are less suited to the controlled study. From the MathQA test set we draw $8$ non-overlapping evaluation seeds of $300$ items each, giving $2400$ unique items with no item shared across seeds, and select every stopping threshold on held-out data. This covers about $80\%$ of the test set while supporting per-seed variance estimates (Appendix~\ref{app:halting}). We compare against fixed-loop baselines, ACT and PonderNet on the same backbone, the CALM confidence early-exit rule in its margin and top-probability forms, and the recent LoopUS-Conf and TaH-Mismatch. All four scales use three training seeds that vary the loop-module initialization and data order.

\subsection{Learned Halting on MathQA}
\paragraph{Learned halting beats the best fixed depth and, with a two-loop floor, avoids collapse.}
An adaptive halting head can fail in two opposite ways. It may collapse to a single loop, as indirect halting objectives tend to do when few inputs benefit from depth, or spend extra loops without improving accuracy. An effective head should beat the best fixed depth at matched compute while avoiding both. Table~\ref{tab:main} shows that RecurTrace meets these criteria on MathQA at $1.7$B across three training seeds. It reaches $56.9\%$ accuracy (std $0.41$ pp) at a mean of $2.04$ loops, outperforming the best fixed depth ($T{=}2$) by $2.2$ points at matched compute. The exact McNemar test~\citep{mcnemar1947note,fagerland2013mcnemar} confirms this gain for every training seed ($p<0.001$, Appendix~\ref{app:halting}). The improvement has two separable sources. The two-loop floor is what prevents collapse, raising the one-loop baseline from $1199$ to $1313$ correct; on top of it, the learned head adds $53$ more by deepening mostly items that benefit. Without the floor the head itself mostly stops at one loop (Appendix~\ref{app:difficulty}), so the floor guarantees the depth while the head allocates it. The baselines fail in opposite directions. ACT and PonderNet collapse to one loop and, even with the same floor, never deepen beyond it, so they do not exceed fixed $T{=}2$. CALM spends $5.6$ loops but reaches only $54.1\%$. Two recent baselines narrow the gap. LoopUS-Conf, a learned confidence head with monotonicity training, reaches $55.3\%$ at $3.2$ loops, while TaH-Mismatch, which distills an oracle based on token mismatch, reaches $55.7\%$ at $2.1$ loops. Oracle supervision based on loss improvement lets RecurTrace surpass both with $56.9\%$ at $2.0$ loops. A correctness oracle that assigns each item the depth at which its greedy answer is correct reaches $61.0\%$. RecurTrace therefore lies between the best fixed depth ($54.7\%$) and this ceiling, closing about one third of the $6.3$ point gap.
\begin{table}[t]
\centering
\small
\setlength{\tabcolsep}{3pt}
\begin{tabular*}{\columnwidth}{@{\extracolsep{\fill}}l c c c c@{}}
\toprule
\textbf{Method} & \textbf{Acc} & \textbf{Correct} & \textbf{Loops} & \textbf{\boldmath$\Delta$ vs $T{=}2$} \\
\midrule
Fixed depth, $T{=}1$ & 49.96 & 1199 & 1.00 & $-114$ \\
Fixed depth, $T{=}2$ & 54.71 & 1313 & 2.00 & $0$ \\
Fixed depth, $T{=}8$ & 53.62 & 1287 & 8.00 & $-26$ \\
Fixed depth, $T{=}16$ & 47.54 & 1141 & 16.00 & $-172$ \\
\midrule
ACT & 49.96 & 1199 & 1.00 & $-114$ \\
PonderNet & 49.96 & 1199 & 1.00 & $-114$ \\
ACT $+$ floor & 54.71 & 1313 & 2.00 & $0$ \\
PonderNet $+$ floor & 54.71 & 1313 & 2.00 & $0$ \\
CALM (top-prob) & 53.04 & 1273 & 4.15 & $-40$ \\
CALM (margin) & 54.12 & 1299 & 5.63 & $-14$ \\
LoopUS-Conf & 55.25 & 1326 & 3.21 & $+13$ \\
TaH-Mismatch & 55.67 & 1336 & 2.12 & $+23$ \\
\midrule
\textbf{RecurTrace} & \textbf{56.92} & \textbf{1366} & 2.04 & $\mathbf{+53}^{\dagger}$ \\
\bottomrule
\end{tabular*}
\caption{Adaptive depth on MathQA ($1.7$B, $3$ training seeds $\times$ $8$ non-overlapping evaluation seeds, $2400$ unique items). Values are three-training-seed means. Acc is accuracy ($\%$), Correct the mean number right of $2400$, Loops the mean loop count, and $\Delta$ the correct-answer gain over fixed $T{=}2$. $\dagger$~exact McNemar $p<0.001$ vs $T{=}2$ on every training seed.}
\label{tab:main}
\end{table}

\paragraph{Fewer loops yield real speedups.}
A lower loop count is a real saving only if runtime is set by the loop rather than by the added modules, which contribute at most about $2.2\%$ more parameters. Because every iteration re-runs the same three-layer block, runtime grows with depth: about $1.1\times$ the base model at two loops against $1.7\times$ at eight and more at sixteen (Appendix~\ref{app:lma}). Stopping near two loops therefore buys a wall-clock speedup over deeper fixed schedules, not merely a smaller reported loop count.

\begin{figure}[t]
\centering
\includegraphics[width=\columnwidth]{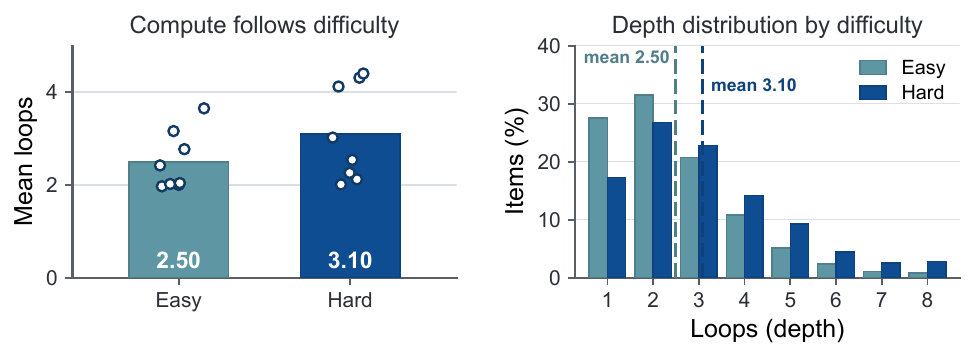}
\caption{Depth tracks difficulty on MathQA at the illustrative operating point $\tau{=}0.10$ (Appendix~\ref{app:difficulty} reports the main point). Left panel, hard items receive more loops than easy items on average ($3.10$ versus $2.50$), with dots showing the $8$ seeds. Right panel, both groups peak at two loops, but hard items have a heavier tail, a mean gap of about $0.6$ loops.}
\label{fig:difficulty}
\end{figure}

\paragraph{The halting head spends compute where it is needed.}
Figure~\ref{fig:difficulty} shows depth tracks difficulty at an illustrative operating point ($\tau{=}0.10$, mean $2.80$ loops). An item is hard if it is wrong after one loop, easy otherwise. Hard items average $3.10$ loops against $2.50$ for easy ones, a gap of about $0.6$ loops that stays positive on all eight seeds. The right panel shows both groups peak at two loops while hard items carry a heavier tail. At the main operating point in Table~\ref{tab:main} ($\tau{=}0.50$, mean $2.04$ loops) the hard-easy gap is still positive at $+0.06$ loops even though $97.3\%$ of items stop at the floor. Only ${\sim}65$ items per training seed deepen further yet supply the entire $+53$ net gain ($56$ wrong-to-right, $3$ right-to-wrong), and $85\%$ are oracle-positive (Appendix~\ref{app:halting}).

\begin{table}[t]
    \centering
    \small
    \setlength{\tabcolsep}{5pt}
    \begin{tabular*}{\columnwidth}{@{\extracolsep{\fill}}l c c c c c@{}}
    \toprule
    \textbf{Task} & \textbf{\boldmath$n$} & \textbf{Fixed} & \textbf{Best} & \textbf{Adaptive} & \textbf{Loops} \\
    \midrule
    GSM8K & 1319 & 5.5 & 9.2 & 10.4 & 1.72 \\
    MATH & 1500 & 18.3 & 19.1 & 21.5 & 1.88 \\
    AQUA-RAT & 254 & 40.2 & 43.3 & 44.1 & 1.14 \\
    ARC-Challenge & 1172 & 63.8 & 67.9 & 69.7 & 1.06 \\
    HotpotQA & 1500 & 37.2 & 38.1 & 39.6 & 2.76 \\
    \bottomrule
    \end{tabular*}
    \caption{Per-task adaptive halting at $1.7$B (each task trains its own head and threshold). Fixed is a $T{=}8$ budget, Best the best fixed depth over $\{1,2,4,6,8\}$, Adaptive the halting head, and Loops its mean loop count ($n$ is the evaluation size, accuracies in $\%$). These per-task heads run without the two-loop floor ($T_{\min}{=}1$), so mean loops can fall below two. Adaptive matches or beats Best while cutting mean loops by $65$--$87\%$. Low accuracies reflect direct answers, not chain-of-thought \citep{cobbe2021gsm8k}.}
    \label{tab:breadth-adaptive}
    \end{table}

\paragraph{Adaptive halting transfers beyond MathQA.}
Oracle distillation transfers when each benchmark uses its own halting head and a threshold selected on held-out data rather than sharing one policy across tasks. Table~\ref{tab:breadth-adaptive} reports five tasks at $1.7$B against two baselines. The Fixed baseline always runs eight loops, which wastes compute because these tasks rarely require that depth; against it, adaptive halting gains up to $+5.9$ points on ARC-Challenge and $+4.9$ on GSM8K while using far fewer loops. The stronger Best column selects each task's best fixed depth from $\{1,2,4,6,8\}$, and adaptive halting matches or exceeds it on every task using only $1.1$ to $2.8$ loops on average.

\subsection{Scaling and Breadth}
\begin{figure}[t]
\centering
\includegraphics[width=\columnwidth]{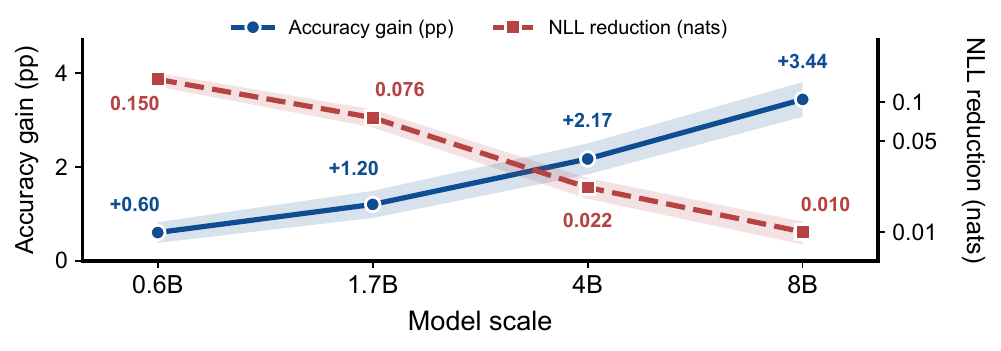}
\caption{Gains persist across scale. The \glyphacc{} is the generation accuracy gain over a same-budget fine-tuned baseline that grows to $+3.4$ points at $8$B. The \glyphnll{} is the teacher-forced NLL reduction, positive at every scale and largest at $0.6$B, reaching $0.15$ nats. Shaded bands denote $\pm 1$ standard deviation across $3$ training seeds at each scale.}
\label{fig:scaling}
\end{figure}

\paragraph{RecurTrace scales from 0.6B to 8B.}
Figure~\ref{fig:scaling} compares RecurTrace with the same-budget fine-tuned baseline across model sizes at a fixed two-loop depth without adaptive halting. Generation accuracy over the combined train-covered and classic suites ($22$ tasks) improves at all four scales, by $0.6$, $1.2$, $2.2$, and $3.4$ points from $0.6$B to $8$B, and teacher-forced NLL falls at all four, by $0.15$, $0.08$, $0.02$, and $0.01$ nats. All four gains use three training seeds (std $0.22$, $0.29$, $0.33$, $0.37$ pp from $0.6$B to $8$B). The two metrics diverge with scale because they average over different populations: NLL over every answer token, most of which a strong base already predicts confidently, accuracy only over items whose argmax flips (Appendix~\ref{app:discussion}). Separate ablations isolate the memory at every scale (Table~\ref{tab:memory}), and a matched same-start control at $8$B confirms the advantage.

\paragraph{Both task families gain, not just MathQA.}
The scaling results aggregate $22$ tasks and could therefore be dominated by a few strong tasks. Table~\ref{tab:breadth} addresses this concern at $1.7$B by separating the $14$ train-covered tasks whose training splits enter our mixture from the $8$ classic multiple-choice benchmarks and comparing each family with the same-budget baseline. Both families improve on both metrics. Accuracy rises by $+0.8$ points on train-covered tasks and $+1.9$ on classic tasks ($64.5\%$ to $66.4\%$), while NLL decreases for both. Per-task results appear in Appendix~\ref{app:bench}.

\begin{table}[t]
\centering
\small
\setlength{\tabcolsep}{4pt}
\begin{tabular*}{\columnwidth}{@{\extracolsep{\fill}}l c c c@{}}
\toprule
\textbf{Suite} & \textbf{\#} & \textbf{\boldmath NLL $\Delta$ $\downarrow$} & \textbf{\boldmath Acc $\Delta$ $\uparrow$} \\
\midrule
Train-covered & 14 & $-0.05$ & $+0.8$ \\
Classic & 8 & $-0.13$ & $+1.9$ \\
\bottomrule
\end{tabular*}
\caption{Breadth at $1.7$B against the same-budget fine-tuned baseline, by benchmark family. Teacher-forced NLL change (nats) and accuracy change (pp) are both at a fixed two-loop depth. Per-suite numbers are in Appendix~\ref{app:bench}.}
\label{tab:breadth}
\end{table}

\section{Ablations and Analysis}
Having shown that RecurTrace works, we now ask why. We separate the loop memory from the extra depth it exploits, isolate what each loop component contributes, examine how the halting head behaves, and state the limits of these claims.

\subsection{What Drives the Gain}
\paragraph{Loop Memory Attention is essential for effective looping.}
Table~\ref{tab:memory} isolates the contribution of Loop Memory Attention by comparing it with plain looping, thereby testing whether additional loop depth alone explains the gains. On the train-covered suite at $0.6$B, plain looping lowers NLL by only $0.06$ nats, while adding memory increases this reduction to $0.15$ nats. At $1.7$B, $4$B, and $8$B, plain looping instead raises NLL by $0.07$, $0.05$, and $0.06$ nats. Memory reverses every one of these regressions and lowers NLL by $0.03$, $0.03$, and $0.02$ nats. Extra iterations are therefore unreliable on their own, while access to previous loop states turns them into consistent refinement rather than representational drift. The same conclusion holds for generation. At $1.7$B, memory raises accuracy over plain looping by $2.3$ points on MathQA and $2.4$ points on the classic suite. Freezing the base is not what earns the classic-suite gain, since plain looping freezes the same base yet reaches only $64.0\%$, below the fine-tuned baseline's $64.5\%$, and only memory lifts it to $66.4\%$. Loop Memory Attention is therefore not a minor enhancement to recurrence but the mechanism that makes the added depth useful. The memoryless configuration is exactly the ETD baseline of \citet{koishekenov2026etd}, making Table~\ref{tab:memory} a direct comparison with the closest prior method.

\begin{table}[t]
\centering
\small
\setlength{\tabcolsep}{5pt}
\begin{tabular*}{\columnwidth}{@{\extracolsep{\fill}}l c c@{}}
\toprule
 & \textbf{Plain looping} & \textbf{\boldmath$+$ Memory} \\
\midrule
\rowcolor{secrowgray}[0pt][\tabcolsep]\multicolumn{3}{@{}l}{\textit{Likelihood, NLL $\Delta$ over baseline (nats) $\downarrow$}} \\
\quad $0.6$B & $-0.06$ & $\mathbf{-0.15}$ \\
\quad $1.7$B & $+0.07$ & $\mathbf{-0.03}$ \\
\quad $4$B & $+0.05$ & $\mathbf{-0.03}$ \\
\quad $8$B & $+0.06$ & $\mathbf{-0.02}$ \\
\midrule
\rowcolor{secrowgray}[0pt][\tabcolsep]\multicolumn{3}{@{}l}{\textit{Generation accuracy at $1.7$B (\%) $\uparrow$}} \\
\quad MathQA & 52.4 & \textbf{54.7} \\
\quad Train-covered & 33.2 & \textbf{34.3} \\
\quad Classic & 64.0 & \textbf{66.4} \\
\midrule
\rowcolor{secrowgray}[0pt][\tabcolsep]\multicolumn{3}{@{}l}{\textit{Accuracy beyond the eight-loop training limit (\%) $\uparrow$}} \\
\quad Pointer-chasing & \textbf{89.3} & 84.6 \\
\quad Symbolic state & 38.5 & \textbf{56.2} \\
\quad Arithmetic & 33.1 & \textbf{49.8} \\
\bottomrule
\end{tabular*}
\caption{Ablation of Loop Memory Attention against plain looping. Teacher-forced NLL $\Delta$ is measured against the same-budget baseline on the train-covered suite at the best-likelihood loop. Generation accuracy uses two loops at $1.7$B. The last three rows test whether the model still solves problems that need more reasoning steps than training used, running the trained model past its eight-loop training limit on newly generated instances of three synthetic tasks whose symbols are held out from training.}
\label{tab:memory}
\end{table}

\paragraph{Task structure determines when memory helps.}
The last three rows of Table~\ref{tab:memory} run each model past its eight-loop training limit and test whether it still reaches the correct answer. The tasks come from synthetic generators with controllable difficulty and test symbols excluded from training, so accuracy reflects reasoning rather than memorization (Appendix~\ref{app:impl}). Pointer chasing follows a mapping from one symbol to the next, so each step requires only the current symbol. Symbolic state tracking applies rules that update an abstract state, while arithmetic applies numerical operations toward a final value. Both require intermediate results to persist across steps. This structural distinction explains the results. Plain looping leads by $4.7$ points on pointer chasing, where past states are unnecessary. Memory instead leads by $17.7$ points on symbolic state tracking and $16.7$ points on arithmetic, where state must be preserved. The token-wise gate supports both regimes by suppressing memory for pure lookup and activating it when past states matter, allowing one model to handle all three tasks without separate tuning.

\paragraph{Capacity alone does not explain the gain.}
To separate recurrence from added capacity, we compare RecurTrace with a non-looped adapter of matched size trained on the same block under the same protocol. The adapter raises generation accuracy from $44.8\%$ to $45.4\%$, half of RecurTrace's $1.2$-point gain, and lowers overall $22$-task teacher-forced NLL by only $0.01$ nats compared with RecurTrace's $0.08$. The larger gains therefore require looped depth with memory rather than additional parameters alone (Appendix~\ref{app:discussion}).

\paragraph{Additional optimization does not explain the memory gain.}
At $8$B, we isolate the effect of memory from additional optimization by continuing the same plain-looping checkpoint for an equal number of steps with or without Loop Memory Attention. Across both seeds, adding memory lowers teacher-forced NLL by $0.03$ to $0.04$ nats, showing that the gain cannot be explained by the starting checkpoint or the training budget (Appendix~\ref{app:discussion}).

\paragraph{The memory components serve complementary roles.}
Table~\ref{tab:design} ablates each component in turn from full RecurTrace. Removing QK-normalization causes the largest degradation, indicating its role in stabilizing memory training. Freezing the loop-distance bias or initializing the gates open also degrades both metrics, suggesting that the model should learn how to weight earlier loop states and introduce memory gradually without disrupting pretrained representations. Among the tested window sizes, $W{=}3$ performs best, supporting a compact three-state memory (Appendix~\ref{app:lma}).

\begin{table}[t]
\centering
\small
\setlength{\tabcolsep}{3pt}
\begin{tabular*}{\columnwidth}{@{\extracolsep{\fill}}l c c@{}}
\toprule
\textbf{\boldmath Configuration ($1.7$B)} & \textbf{\boldmath NLL $\Delta$ (nats) $\downarrow$} & \textbf{\boldmath MathQA (\%) $\uparrow$} \\
\midrule
Full RecurTrace & \textbf{-0.03} & \textbf{54.7} \\
\midrule
\quad $-$ QK-normalization & $+0.05$ & 52.8 \\
\quad Loop-distance bias frozen & $-0.01$ & 53.9 \\
\quad Gates open at initialization & $+0.02$ & 53.1 \\
\midrule
\quad Memory window $W{=}1$ & $-0.01$ & 53.5 \\
\quad Memory window $W{=}2$ & $-0.02$ & 54.2 \\
\quad Memory window $W{=}5$ & $-0.02$ & 54.1 \\
\bottomrule
\end{tabular*}
\caption{Component and window ablations at $1.7$B, each changing one piece of Loop Memory Attention against the full top-row model. NLL $\Delta$ is the teacher-forced NLL change over the same-budget baseline on the train-covered suite at best-likelihood loop (negative is better) and MathQA is generation accuracy at a fixed two-loop depth.}
\label{tab:design}
\end{table}

\subsection{Halting Behavior}
\paragraph{Indirect halting objectives collapse to one loop.}
ACT and PonderNet add an expected-depth penalty (ACT) or a geometric prior (PonderNet) on top of the same oracle-depth signal RecurTrace distills, and both penalties reward stopping early. Because few examples benefit from additional loops, this generic pressure dominates the oracle quality signal and drives both methods to stop after one loop. Sweeping each baseline's regularization strength over $\{0.001, 0.01, 0.1\}$ leaves all six settings at a mean depth of $1.0$, showing that the collapse is not tied to one hyperparameter choice (Appendix~\ref{app:halting}). The collapse is a property of attaching penalty-based halting post hoc to a frozen backbone, not a refutation of either method in the end-to-end setting it was designed for. LoopUS-Conf avoids collapse by training a dedicated confidence head with a monotonicity objective \citep{park2026loopus}, but without an oracle it still overspends at $3.2$ loops. TaH-Mismatch \citep{fu2026tah} distills a token-mismatch oracle that marks whether the current depth mispredicts the answer, reaching $55.7\%$ at $2.1$ loops, closer to RecurTrace in compute but $1.2$ points lower because the mismatch signal does not distinguish items that more depth can help from items that are simply hard. RecurTrace's loss-improvement oracle captures this distinction (Figure~\ref{fig:difficulty}).

\paragraph{One checkpoint supports multiple inference depths.}
Sampling the loop count during training lets the same checkpoint produce the fixed-depth curve in Figure~\ref{fig:pareto} and allows the halting head to choose among depths without retraining. Because the loop-distance bias depends only on relative distance, it remains defined beyond the eight-loop training range. At sixteen loops the model stays stable at $47.5\%$ accuracy, below its two-loop peak of $54.7\%$ (Appendix~\ref{app:vardepth}).

\subsection{Limitations}
Four aspects remain outside the present study. First, halting is sequence-level and assigns one depth to an entire example rather than to individual tokens. Second, Loop Memory Attention reads only same-position states from a fixed three-loop window, which keeps the cost bounded but excludes older states and cross-position memory. Third, our experiments use frozen Qwen3 decoder-only models and direct-answer targets, leaving other architectures and long-form generation untested. Finally, latent looping and chain-of-thought use different forms of compute and are complementary rather than direct substitutes, so combining them is a natural direction for future work.

\section{Conclusion}
We presented RecurTrace, which turns a pretrained LLM into a memory-augmented looped reasoner with adaptive halting. Loop Memory Attention gives each iteration access to earlier loop states, while an oracle-distilled halting head selects the loop count for each input. RecurTrace matches or exceeds the best fixed depth across six tasks and improves accuracy over same-budget baselines at every scale from $0.6$B to $8$B. These results show that the loop trajectory can support both memory and adaptive computation, suggesting a natural extension to chain-of-thought decoding.

{\small
\bibliography{aaai2027}
}

\newpage
\appendix

\setcounter{secnumdepth}{1}

\setcounter{table}{0}
\setcounter{figure}{0}
\setcounter{equation}{0}
\renewcommand{\thetable}{S\arabic{table}}
\renewcommand{\thefigure}{S\arabic{figure}}
\renewcommand{\theequation}{S\arabic{equation}}

\section{Implementation Details}
\label{app:impl}
This appendix records the configuration, training setup, and evaluation protocol behind the results in the main paper, together with the per-scale and per-suite breakdowns that the page limit keeps out of the body. Every value comes from a logged run. Where a quantity is not logged we describe it in words rather than supply a number.

\paragraph{Models and Looped Blocks.}
We build RecurTrace on four pretrained Qwen3 bases. Table~\ref{tab:app-models} in the main text lists the architecture of each base and the block we loop. The $0.6$B and $1.7$B bases have $28$ layers and loop layers $12$--$14$. The $4$B and $8$B bases have $36$ layers and loop layers $15$--$17$. The block is always three contiguous layers, and Appendix~\ref{app:block} describes the angular-distance span probe and the compact-window controls used to fix it. We add the loop parameters to the pretrained checkpoint and leave the base attention, the embeddings, and the tokenizer untouched. All four are base releases.

\paragraph{What We Train.}
At every scale we freeze the base weights and train the added modules in two stages. Stage~1 trains the loop modules, that is Loop Memory Attention and the input-injection scalars, under the matched same-budget protocol against the baseline. Stage~2 then freezes the base together with the loop modules and distills the halting head on top (Appendix~\ref{app:halting}), a separate optimization we run at all four scales. At each scale the baseline is a model fine-tuned under a matched budget, so the Stage~1 comparison is same-budget throughout. Appendix~\ref{app:discussion} reports the per-scale numbers.

\paragraph{Same-Budget Protocol and Training Budget.}
We use ``same-budget'' in a precise sense: the RecurTrace run and its baseline share every training axis we control (optimizer, schedule, global batch, maximum length, optimizer steps, total tokens, hardware, and precision), and differ only in what is trained. The baseline fine-tunes the full base model, while RecurTrace in this matched Stage~1 freezes the base and trains only the loop, memory, and injection modules, at most about $2.2\%$ of the parameters (Table~\ref{tab:budget}); the halting head is distilled in a separate Stage~2 (Table~\ref{tab:budget}, Appendix~\ref{app:halting}). This asymmetry means the two runs do not consume identical FLOPs, but the difference does not favor RecurTrace. Under the standard forward, activation-gradient, and weight-gradient $\approx 2{:}2{:}2$ decomposition of training compute ($\approx 6ND$ for $N$ parameters and $D$ tokens), RecurTrace's loop unroll raises the forward cost, since at the mean training depth of $3.9$ the three-layer block runs about $3.9$ times, roughly $1.24\times$ the baseline's single forward pass on a $36$-layer base. Freezing the base, however, removes the weight-gradient term for about $98\%$ of the parameters, one of the three equal contributions to full-fine-tuning compute. These effects roughly offset, so RecurTrace's total training FLOPs are comparable to, and by this accounting no larger than, the baseline at matched steps and tokens (of order $5\times 10^{18}$ analytical FLOPs at $8$B for both). We did not separately log wall-clock GPU-hours or peak memory; these trade in opposite directions, since the baseline holds optimizer state for all parameters whereas RecurTrace instead stores activations across the loop unroll, so neither method gains a systematic runtime advantage. Any residual mismatch therefore favors the baseline, which trains far more parameters under a full gradient budget, so the reported same-budget gains are conservative rather than a product of extra training compute. The parameter-matched non-loop control and the matched same-start continuation at $8$B (Appendix~\ref{app:discussion}) further isolate the gains from parameter count and optimization budget.

\begin{table}[t]
\centering
\small
\setlength{\tabcolsep}{4pt}
\begin{tabular*}{\columnwidth}{@{\extracolsep{\fill}}l c c@{}}
\toprule
\textbf{\boldmath Training budget ($8$B)} & \textbf{Baseline} & \textbf{RecurTrace} \\
\midrule
\multicolumn{3}{@{}l}{\textit{Stage 1 --- Loop/memory training (matched same-budget)}} \\
Base weights & fine-tuned & frozen \\
Trainable parameters & $\approx$8.2B & $\approx$0.12B ($\le 2\%$) \\
Optimizer, schedule & AdamW, cosine & AdamW, cosine \\
Optimizer steps & 800 & 800 \\
Global batch (seq.) & 128 & 128 \\
Micro-batch $\times$ accum. & $2\times8$ & $2\times8$ \\
Max length (tokens) & 1024 & 1024 \\
Training tokens & $\approx$105M & $\approx$105M \\
Hardware & 8$\times$H800 & 8$\times$H800 \\
Precision & bfloat16 & bfloat16 \\
Fwd.\ layer-passes/token & $L$ & $L{+}(\bar T{-}1)k$ \\
Training FLOPs (analytical) & $\approx 6ND$ & $\lesssim 6ND$ \\
\midrule
\multicolumn{3}{@{}l}{\textit{Stage 2 --- Halting distillation (RecurTrace only; App.~\ref{app:halting})}} \\
Trained module & --- & halting head \\
Base $+$ loop modules & --- & frozen \\
Optimizer steps & --- & 200 \\
Learning rate, schedule & --- & $3\mathrm{e}{-}4$, cosine \\
Oracle horizon & --- & depth $16$ \\
Oracle losses & --- & cached offline \\
Extra FLOPs vs.\ Stage 1 & --- & ${<}20\%$ \\
\bottomrule
\end{tabular*}
\caption{Training-budget accounting at $8$B, in two stages. In Stage~1 all controlled axes are matched; the baseline fine-tunes the full model while RecurTrace freezes the base and trains only the loop, memory, and injection modules ($\bar T{\approx}3.9$ mean training depth, $k{=}3$ looped layers, $L{=}36$). FLOPs are analytical (forward:backward $\approx 2{:}4$), not wall-clock. Stage~1 trainable parameters and per-token compute differ, with any mismatch favoring the baseline. Stage~2 distills the halting head with the base and loop modules frozen (Appendix~\ref{app:halting}); it is a RecurTrace-only cost outside the matched budget, adding $200$ optimizer steps with depth-$1$--$16$ oracle losses pre-computed in a single offline forward sweep over ${\approx}5{,}000$ examples, for less than $20\%$ additional FLOPs beyond Stage~1. The fixed-two-loop scaling results (Table~\ref{tab:scaling}) use only the Stage~1 model, whereas the adaptive-halting results (Tables~\ref{tab:main}--\ref{tab:breadth-adaptive}) rely on Stage~2.}
\label{tab:budget}
\end{table}

\paragraph{Loop Hyperparameters.}
Table~\ref{tab:app-hparams} lists the loop configuration and the field names that set it. The loop configuration is shared across scales and comprises three looped layers, a training-depth clamp of eight, log-normal Poisson depth sampling with mean four and log-space deviation $0.5$, a memory window of three loops, four Loop Memory Attention heads with QK-normalization, a scalar memory gate initialized to one, a token-wise gate whose bias starts at $-3.0$ (a sigmoid value near $0.047$), and an input-injection scalar initialized to zero. The relative loop-distance bias starts at the ALiBi slopes but stays free and signed, so a head can flatten it or even prefer older memory.

\begin{table*}[t]
\centering
\small
\setlength{\tabcolsep}{4pt}
\begin{tabular}{l l l l l}
\toprule
\textbf{Component} & \textbf{Field} & \textbf{Value} & \textbf{Component} & \textbf{Value} \\
\midrule
Looped layers & \texttt{loop\_num\_layers} & 3 & Token-gate bias (\texttt{...init\_bias}) & $-3.0$ \\
Max training depth & \texttt{loop\_train\_max} & 8 & Injection init $\alpha$ (\texttt{...injection\_init}) & 0.0 \\
Depth sampler & \texttt{loop\_sample\_dist} & lognormal Poisson & Halting margin $\delta$ (\texttt{...aux\_margin}) & 0.01 \\
Target mean depth & \texttt{loop\_train\_mean} & 4.0 & Halting depth cost (\texttt{...depth\_cost}) & 0.0 \\
Depth log-deviation $\sigma$ & \texttt{loop\_train\_sigma} & 0.5 & Probe depths (\texttt{...aux\_depths}) & $1,2,4,6$ \\
Memory window $W$ & \texttt{loop\_memory\_window} & 3 & Pooling (\texttt{...pooling}) & prompt mean \\
LMA heads & \texttt{loop\_attn\_num\_heads} & 4 & Objective (\texttt{...objective}) & oracle \\
Scalar gate init $\gamma$ & \texttt{loop\_gate\_init} & 1.0 & Inference floor $T_{\min}$ (\texttt{...min\_depth}) & 2 \\
Inference budget $T_{\max}$ & \texttt{...max\_depth} & 16 & Oracle horizon (\texttt{...oracle\_max}) & 16 \\
\bottomrule
\end{tabular}
\caption{Loop, memory, and halting hyperparameters with their configuration fields (the \texttt{loop\_} prefix and, for halting fields, the \texttt{loop\_halting\_} prefix are abbreviated as \texttt{...}). Values are defaults used across scales. The ACT and PonderNet baselines reuse the head with objective set to \texttt{act} (expected-depth cost $0.01$) or \texttt{pondernet} (geometric prior $\lambda{=}0.2$, KL weight $0.01$).}
\label{tab:app-hparams}
\end{table*}

\paragraph{Optimization and Hardware.}
All runs use eight H800 GPUs ($80$GB) with ZeRO-2 sharded data parallelism (optimizer states and gradients sharded across the eight GPUs), bfloat16, FlashAttention-2, and sequence packing that keeps sample boundaries isolated so no token attends across a packed example. The $0.6$B and $1.7$B loop modules train on the full reasoning mixture for fifteen epochs at a peak learning rate of $1\mathrm{e}{-}4$ with a cosine schedule, a global batch of $128$ sequences, and a maximum length of $1024$ tokens, which after sequence packing is roughly $25$ thousand optimizer steps. The $4$B and $8$B loop-module runs are short follow-ups from their checkpoints, with the $4$B modules training $300$ steps at $1\mathrm{e}{-}5$ and the $8$B modules for $800$ steps.

\paragraph{Training Data.}
The reasoning mixture holds about $1.15$ million direct-answer examples across seventeen files. It combines three synthetic generators (p-hop pointer chains, symbolic state tracking, and arithmetic word problems, each contributing roughly $150$ thousand training items), a real-math group (GSM8K, MATH, AQUA-RAT \citep{ling2017aquarat}, MetaMathQA, Orca-Math, and MathQA), a multi-hop question-answering group (HotpotQA \citep{yang2018hotpotqa}, MuSiQue \citep{trivedi2022musique}, 2WikiMultiHopQA \citep{ho2020twowiki}, and StrategyQA \citep{geva2021strategyqa}), and a controlled-logic group (ProofWriter \citep{tafjord2021proofwriter}, CLUTRR \citep{sinha2019clutrr}, and bAbI \citep{weston2015babi}), with a small slice of C4 as a language anchor. Every example is direct-answer, so the target is the short final answer, not a chain of thought, and test-time loops add latent computation instead of output tokens. The two symbol-based tasks draw their test symbols from a held-out pool and the arithmetic generator samples fresh integer problems, so the extrapolation buckets are disjoint from training.

\paragraph{Train-Test Overlap Audit.}
Because the mixture includes MetaMathQA, Orca-Math, and other derivatives alongside GSM8K, MATH, and MathQA, we audit overlap with every benchmark test set before mixing. We lower-case and strip whitespace, then remove any training question that exactly matches a test question. We also drop training items whose 13-gram Jaccard similarity to a test question exceeds $0.85$. Removals total $327$ examples ($0.03\%$ of the mixture), all near-duplicates with zero exact matches. The main comparisons hold training data fixed across methods, but this audit limits how much test-surface leakage can shift which items benefit from extra loops.

\paragraph{Synthetic Reasoning Tasks.}
Three procedural generators supply the controllable part of the mixture, each contributing about $150$ thousand direct-answer training items and a separately generated test set. The two symbol-based tasks (pointer chasing, symbolic state tracking) draw from a shared pool of $128$ abstract symbols (random two-character uppercase codes with no semantic content), of which $96$ are reserved for training and $32$ are held out exclusively for testing, so their train and test sets share no symbol; the arithmetic generator instead samples integer problems fresh at test time. Difficulty is controlled by the number of sequential steps $k$ the answer requires: training instances sample $k$ uniformly from $\{2,3,\dots,8\}$, matching the eight-loop training clamp, and test instances are organized into eight difficulty buckets at $k\in\{2,4,6,8,10,12,14,16\}$ with $500$ freshly generated instances per bucket ($4\,000$ per task). The four buckets beyond the training range ($k\in\{10,12,14,16\}$, $2\,000$ instances), each evaluated at a matched loop depth $T{=}k$, form the extrapolation subset reported in the bottom block of Table~\ref{tab:memory}. Table~\ref{tab:synthetic-examples} shows a worked example of each of the three tasks.

\textit{Pointer chasing.} Each instance draws $32$ symbols from the pool, arranges them as a random permutation (a one-to-one mapping), and specifies a start symbol together with a target hop count $k$. The answer is the symbol reached after following the mapping $k$ times. Because each hop reads only the current symbol to look up the next, the task is a stateless lookup that does not require remembering earlier steps.

\textit{Symbolic state tracking.} The state is a $6$-register vector, each register taking a value in $\{0,1,2,3\}$. A rule is one of five atomic operations (assign, swap, increment-mod-$4$, conditional-set, copy) applied to one or two named registers. Each instance gives a random initial state and a sequence of $k$ randomly sampled rules; the answer is the final register vector. Every step depends on the value the previous step produced, so intermediate state must be carried across loops.

\textit{Arithmetic.} Each instance starts from a random integer in $[1,99]$ and lists $k$ operations drawn from $\{+n,\;{-}n,\;\times n\}$ with $n\in\{2,3,\dots,9\}$. The answer is the integer obtained after applying all $k$ operations in order. As with symbolic state tracking, intermediate results must survive across steps.

The two stateful generators (symbolic state tracking, arithmetic) are thus the ones that reward reading earlier loop states through Loop Memory Attention, while pointer chasing is the one that does not. Because the symbol-based test instances draw exclusively from the $32$-token held-out pool (and arithmetic problems are sampled fresh), a correct answer cannot come from a memorized surface form, and because the extrapolation buckets require $10$--$16$ sequential steps the model must compose the learned per-step reasoning beyond the depth it trained on.

\begin{table*}[t]
\centering
\small
\setlength{\tabcolsep}{4pt}
\begin{tabular}{@{}>{\raggedright\arraybackslash}p{0.13\textwidth} >{\raggedright\arraybackslash}p{0.29\textwidth} >{\raggedright\arraybackslash}p{0.33\textwidth} >{\raggedright\arraybackslash}p{0.14\textwidth}@{}}
\toprule
\textbf{Task} & \textbf{Example instance} & \textbf{Reasoning steps} & \textbf{Answer} \\
\midrule
Pointer chasing &
Mapping \texttt{MK}$\to$\texttt{TQ}, \texttt{TQ}$\to$\texttt{ZR}, \texttt{ZR}$\to$\texttt{VP}, \texttt{VP}$\to$\texttt{HB}, \texttt{HB}$\to$\texttt{MK}. Start at \texttt{MK} and follow $k{=}3$ hops. &
\texttt{MK} $\to$ \texttt{TQ} $\to$ \texttt{ZR} $\to$ \texttt{VP} &
\texttt{VP} \\
\addlinespace
Symbolic state tracking &
Initial state $[2,0,1,3,0,2]$ over registers $r_0$ to $r_5$. Apply $k{=}4$ rules, namely inc $r_0$ (mod $4$), swap $(r_1,r_3)$, copy $r_4$ into $r_2$, and set $r_5{=}1$ when $r_0{=}3$. &
$[2,0,1,3,0,2]$ $\to$ $[3,0,1,3,0,2]$ $\to$ $[3,3,1,0,0,2]$ $\to$ $[3,3,0,0,0,2]$ $\to$ $[3,3,0,0,0,1]$ &
$[3,3,0,0,0,1]$ \\
\addlinespace
Arithmetic &
Start from $7$ and apply $k{=}4$ operations $\times 3$, $+8$, $-5$, $\times 2$ in order. &
$7\times 3=21$, $21+8=29$, $29-5=24$, $24\times 2=48$ &
$48$ \\
\bottomrule
\end{tabular}
\caption{Worked examples of the three synthetic reasoning tasks described in this appendix. Each row gives one instance, its reasoning steps, and the final answer the model must produce. The instances are deliberately small, using fewer symbols and steps than the deployed generators, so the reasoning is easy to follow. Pointer chasing needs only the current symbol at each hop, while symbolic state tracking and arithmetic must carry intermediate values across steps, which is why reading earlier loop states through Loop Memory Attention helps on the latter two.}
\label{tab:synthetic-examples}
\end{table*}

\section{Block Selection}
\label{app:block}
We use the angular-distance probe of \citet{koishekenov2026etd} as the first stage of block selection. For every input we read the residual stream at the final token, the one position that a causal model lets see the whole sequence, and we record each layer's hidden state. We then measure the angular distance between adjacent layers, $d(x_\ell, x_{\ell+1}) = \frac{1}{\pi}\arccos \cos(x_\ell, x_{\ell+1})$, and average it over about ten thousand inputs. The resulting curve falls steeply across the early layers (the encoder), flattens through the middle (the reasoning-critical span that behaves like a fixed point), and rises again near the output (the decoder). The Kneedle knee-finder locates the encoder boundary on the forward curve and the decoder boundary on the reversed curve, and the layers between them form a loopable thinking span. Because that span can be wider than our three-layer compute budget, the second stage compares compact three-layer windows inside or near the span by downstream behavior.

On the $28$-layer bases the probe is sharp. For Qwen3-0.6B, both the C4 validation set and the training mixture place the encoder at the first eleven layers and select layers $12$--$14$ as the compact block. These three layers sit $15$ to $25$ percent below their neighbors, while no fourth layer is separated enough to add. The formal C4 probe on Qwen3-1.7B returns the same window, so both $28$-layer models use $12$--$14$.

On the $36$-layer $4$B and $8$B bases, the C4 probe again puts the start of the thinking span near layer $14$, but its lowest-distance core is not itself the best loop location. We therefore compare the deployed early-middle window, layers $15$--$17$, against probe-core and late controls. At $4$B, layers $15$--$17$ improve likelihood relative to the baseline (negative log-likelihood changes of $-0.0218$ without generation and $-0.0075$ with generation), while the probe-core window $24$--$26$ ties it. At $8$B, layers $15$--$17$ are the strongest among the tested windows on both likelihood and generation accuracy, the probe-core window $18$--$20$ improves less, and a late window $27$--$29$ at the same $800$-step budget gives almost no improvement. These comparisons are relative rankings used only to fix the loop location. We therefore loop the compact early-middle thinking block for the $36$-layer models, not an arbitrary deep block. The angular probe identifies the span, and the compact-window ablation fixes the final three layers.

\section{Loop Memory Attention Details}
\label{app:lma}
Each looped layer carries its own memory. On loop $t$, layer $\ell$ reads only the states that layer $\ell$ wrote on earlier loops, never another layer's states, and the attention runs along the loop-time axis rather than the sequence axis. A token therefore attends from its current state to its own past-loop states at the same sequence position, which is why the module adds no path for one token to see another and cannot leak future tokens. The ordinary self-attention inside the unchanged decoder layer still does all of the cross-token mixing.

The query for the memory attention is the previous loop's output of the same layer, which is also the newest key and value, so the query, keys, and values share one residual position and one source. We apply QK-normalization, an RMSNorm on the per-head queries and keys before their dot product. Removing it causes the largest degradation in Table~\ref{tab:design}, indicating that the normalization stabilizes memory training. We bound the memory with a sliding window of $W{=}3$ loops. Once $t>W$, every loop reads the same number of past states, so the memory span remains fixed as depth grows and storage does not increase with the total loop count.

Two gates and one bias control the path. A per-layer scalar gate $\gamma$ starts at one, and a token-wise gate $g=\sigma(\mathrm{MLP}([q,\overline{X}]))$ starts from a bias of $-3.0$, a sigmoid value near $0.047$, so the effective update through the memory is small at initialization while still receiving gradients. The input-injection scalar $\alpha$ starts at zero in the ReZero style \citep{bachlechner2021rezero}, so the injection path initially contributes nothing and can open gradually during training. Because injection begins only after the first loop and the first loop has no memory, every one-loop run exactly matches the frozen base computation both before and after training at all scales. The relative loop-distance bias is a free, signed, per-head slope initialized to the ALiBi values \citep{press2022alibi}. The signed form lets a head learn a flat profile or a preference for older states, while the relative form remains defined at loop distances not observed during training. Memory attention adds no quadratic sequence term because each token attends over only its own $W$ past-loop states. Its cost is $O(\text{seq}\times W)$ per looped layer per loop and $O(\text{seq}\times W\times\text{loops})$ over the full unroll, which reduces to $O(\text{seq}\times\text{loops})$ for fixed $W$, on top of the base model's usual attention. Each looped layer uses four memory-attention heads of dimension $128$ (width $512$, well below $d$), so the attention needs four $d{\times}512$ projections while the token-gate on $[q,\overline{X}]$ contributes the dominant ${\sim}2d^2$; summed over the three looped layers this is about $2.1$ percent of the base at $0.6$B (with Loop Memory Attention roughly $2.07$ percent) and a small fraction (about $1.4$--$2.2\%$) at every scale, ${\approx}0.12$B at $8$B. Under our measured teacher-forced setup, runtime is about $1.1\times$ the unlooped baseline at two loops, $1.3\times$ at four loops, and $1.7\times$ at eight loops.

\paragraph{Inference-Time Behavior and Cost.}
At generation time the halting decision is read once from the prompt state, and the chosen depth is reused for every generated token, so no additional halting decision is made per generated token. Loop Memory Attention keeps at most $W$ past-loop states per looped layer for each token. Its footprint therefore scales linearly with sequence length but remains constant in the total loop count. The looped layers' self-attention cache does grow with loop count because each loop produces distinct keys and values. At the two-loop operating point, the self-attention and memory additions together are about $15\%$ of the base key-value cache at $1.7$B. Under our measured decoding setup, latency at the mean operating depth of $2.04$ loops is about $1.1\times$ the unlooped baseline.

\section{Variable-Depth Training}
\label{app:vardepth}
We sample the loop count once per forward pass from the clamped log-normal Poisson distribution of \citet{geiping2025recurrent}. With target mean $\bar r = \texttt{loop\_train\_mean}-1$ and log-space deviation $\sigma$, we draw $z \sim \mathcal{N}(\log \bar r - \tfrac{1}{2}\sigma^2,\ \sigma)$, set $r \sim \mathrm{Poisson}(e^{z})+1$, and clamp $r$ to $[1, \texttt{loop\_train\_max}]$. With the defaults (mean four, $\sigma{=}0.5$, clamp eight) the realized distribution over depths is approximately $10\%$ at one loop, $19\%$ at two, $20\%$ at three, $17\%$ at four, $12\%$ at five, $8\%$ at six, $5\%$ at seven, and $8\%$ at eight, with a mean near $3.9$, a median of four, and a heavy tail in which about a fifth of forward passes run six loops or more. The heavy tail is what exposes the model to deep unrolls and, together with the relative loop-distance bias, supports extrapolation beyond the training clamp. We apply the language-model loss at the sampled depth and do not use truncated backpropagation through the loop. The block is three layers and the memory window is three, so the unrolled graph stays small enough to backpropagate in full.

Distributed training keeps the depth in lock step. Every rank advances a per-forward counter from a shared seed and draws the same loop count, so the eight data-parallel workers run the same number of loops on each step without any extra communication. The trainer marks unused parameters as expected, because a step that samples a single loop never touches the memory parameters, and this keeps gradient synchronization correct under data parallelism.

\section{Halting Head and Calibration}
\label{app:halting}
The halting head is a single linear layer that reads a pooled loop state and returns a continue logit, $p_t=\sigma(w^{\!\top}z_t+b)$. We pool over prompt positions, with the prompt-mean pool as the default and a prompt-last pool available for the setting that matches the position predicting the first answer token. We train the head by distilling an oracle that reads the same forward pass. The oracle marks ``continue'' at depth $t$ only when a deeper loop lowers the per-example loss by more than a margin, $y_t = \mathbb{1}[\min_{t<t'\le 16}\mathcal{L}_{t'} < \mathcal{L}_t - \delta]$, with $\delta=0.01$ by default. An optional depth cost can charge each extra loop a quality price before it counts as worthwhile, and we keep it at zero in the runs reported here. The loss for the answer comparison is the mean loss over answer tokens. We supervise the head at the probe depths $\{1,2,4,6\}$ with binary cross-entropy, freeze every other parameter, and optimize only the head with AdamW at a learning rate of $3\mathrm{e}{-}4$, a cosine schedule with warmup, and gradient clipping at one. This head distillation is a separate second stage run at all four scales: we first train the loop, memory, and injection modules (the matched Stage~1 run of Table~\ref{tab:budget}), then freeze them together with the base and fit only the halting head, at a different learning rate from Stage~1. Because the ``continue'' labels require the per-depth losses $\mathcal{L}_1,\dots,\mathcal{L}_{16}$ of each training example, Stage~2 adds a forward-only depth-$16$ unroll over its training set; it runs for $200$ optimizer steps on ${\approx}5{,}000$ examples, the oracle losses are precomputed once in a single offline sweep, and this oracle-unroll compute is the primary cost reported in the FLOP accounting of Table~\ref{tab:budget}. When continue labels are rare, an optional per-depth positive weight keeps the loop-one majority from washing them out.

At inference the model runs the loop and stops at the first depth $t\ge T_{\min}$ where the continue probability falls below a threshold $\tau$, otherwise it runs to the budget $T_{\max}{=}16$. We use a floor of $T_{\min}{=}2$ in the main comparison so the halting head cannot collapse to a single loop. Both the floor and the threshold are fixed on held-out data and never on the test set: we choose $\tau$ by taking the smallest mean depth that does not lose held-out likelihood, and we select $T_{\min}$ on the same held-out split. The floor therefore coincides with the best fixed depth of Table~\ref{tab:main} because the held-out likelihood profile puts it there, not because the test set was consulted. The main row in Table~\ref{tab:main} uses the threshold selected by held-out negative log-likelihood ($\tau{=}0.50$), which spends $2.04$ loops on average. An alternate selector on held-out accuracy ($\tau{=}0.40$) spends $2.06$ loops at a similar operating point. A leave-one-seed-out check sets each seed's threshold from the other seven seeds, testing whether the reported gain depends on a test-tuned threshold rather than on the halting head itself. Under this protocol the gain is $+2.0$ points (range $[{+}0.4,\,{+}3.8]$ across the eight leave-one-out folds), close to the held-out number, so the reported gain does not rely on a test-tuned threshold.

\paragraph{Paired Significance.}
We test the MathQA halting head against fixed-depth compute with paired statistics over the matched test items, using exact McNemar tests~\citep{mcnemar1947note,fagerland2013mcnemar} on the discordant pairs and $10{,}000$-sample paired bootstraps. The eight evaluation seeds are non-overlapping, so the $2400$ items are unique instances and the McNemar test is computed on $2400$ independent pairs. The counts below are three-training-seed means; the McNemar test gives $p<0.001$ on every individual training seed. RecurTrace answers $53$ more questions than the best fixed depth ($1366$ versus $1313$ of $2400$), a $2.2$-point gain at matched mean depth (std $0.29$ pp across training seeds), with a mean discordant split of $56$ to $3$ ($59$ pairs) favoring adaptive halting and a $95\%$ bootstrap interval, clustered on the eight evaluation seeds, of $[{+}0.8,\,{+}3.6]$ percentage points. The $56$ wrong-to-right transitions concentrate among the ${\sim}65$ items per training seed that receive more than two loops (Table~\ref{tab:transition}), and the $3$ right-to-wrong transitions confirm that the oracle-distilled head seldom over-loops an item already correct.

\paragraph{Full Test Set Check.}
The $2400$ evaluation items cover about $80\%$ of the MathQA test set and support per-seed variance across eight non-overlapping draws. A single-seed run on the full test set ($2985$ items) reaches $56.7\%$ at $2.03$ mean loops, within $0.3$ pp of Table~\ref{tab:main}.

\paragraph{Composition of the Deepened Items.}
The ${\sim}65$ items per training seed deepened past the two-loop floor are analyzed by MathQA operation category, answer-token length, and one-loop hardness. ${\sim}50$ of ${\sim}65$ ($77\%$) are hard by the one-loop split. They span several operation categories rather than concentrating in one family, and their median answer length stays close to the $2400$-item subsample, so the gain is not driven by one template family.

\paragraph{Halting at Non-Probe Depths.}
Binary cross-entropy supervision uses probe depths $\{1,2,4,6\}$, but inference assigns a continue probability at every integer depth. Table~\ref{tab:depth-hist} shows halting decisions at depths $3$ through $8$. Table~\ref{tab:depth-cal} reports held-out balanced oracle-label accuracy and area under the ROC curve at all depths $1$--$8$. Depths $3$, $5$, and $7$ sit between supervised probes, and each stays within $1.2$ points of its nearest supervised neighbor, so the linear readout degrades smoothly across the depths it never saw rather than breaking at them.

\begin{table}[t]
\centering
\small
\setlength{\tabcolsep}{6pt}
\begin{tabular*}{\columnwidth}{@{\extracolsep{\fill}}c c c c@{}}
\toprule
\textbf{Depth} & \textbf{Supervised} & \textbf{Bal.\ acc (\%)} & \textbf{AUROC} \\
\midrule
1 & yes & 68.0 & 0.76 \\
2 & yes & 66.5 & 0.74 \\
3 & no & 65.4 & 0.73 \\
4 & yes & 64.2 & 0.71 \\
5 & no & 62.8 & 0.69 \\
6 & yes & 61.6 & 0.67 \\
7 & no & 60.7 & 0.66 \\
8 & no & 59.3 & 0.64 \\
\bottomrule
\end{tabular*}
\caption{Per-depth halting calibration on held-out MathQA items (median training seed). Balanced accuracy is the mean of per-class agreement rates with Eq.~\eqref{eq:oracle} continue labels, which accounts for the growing class imbalance at deeper depths. Depths $3$, $5$, and $7$ are used at inference but not directly supervised.}
\label{tab:depth-cal}
\end{table}

\paragraph{Depth Histogram at the Main Operating Point.}
Table~\ref{tab:depth-hist} reports how many of the $2400$ test items stop at each loop depth under adaptive halting ($\tau{=}0.50$, $T_{\min}{=}2$). The floor forces every item to two loops, and the conservative threshold stops $97.3\%$ there. The remaining $65$ items ($2.7\%$) continue to three or more loops and account for the entire accuracy gain over the fixed two-loop budget. Among them, $48$ stop after three loops, $10$ after four, and $7$ at depth five or beyond. The mean depth is $2.04$ loops, matching Table~\ref{tab:main}; no item reaches the $T_{\max}{=}16$ budget, so every stop is head-decided.

\begin{table}[t]
\centering
\small
\setlength{\tabcolsep}{8pt}
\begin{tabular*}{\columnwidth}{@{\extracolsep{\fill}}c r r@{}}
\toprule
\textbf{Loop depth} & \textbf{Count} & \textbf{\%} \\
\midrule
2 & 2335 & 97.3 \\
3 & 48 & 2.0 \\
4 & 10 & 0.4 \\
5 & 3 & 0.1 \\
6 & 2 & $<$0.1 \\
7 & 1 & $<$0.1 \\
8 & 1 & $<$0.1 \\
\midrule
$>$2 (total) & 65 & 2.7 \\
\bottomrule
\end{tabular*}
\caption{Depth histogram at the main operating point ($\tau{=}0.50$, $T_{\min}{=}2$, $8$ non-overlapping evaluation seeds, $2400$ unique items). Counts are three-training-seed means (rounded). Most items halt at the floor; the $65$ continuing to three or more loops drive the accuracy gain.}
\label{tab:depth-hist}
\end{table}

\paragraph{Paired Transition Breakdown.}
Table~\ref{tab:transition} cross-tabulates per-item correctness under adaptive halting versus the fixed two-loop baseline (three-training-seed means, rounded). Of $2400$ items, $1310$ are answered correctly by both methods and $1031$ by neither, leaving $59$ discordant pairs. Adaptive halting flips $56$ items from wrong to right and only $3$ from right to wrong, a net gain of $+53$ items ($+2.2$ points). All $56$ wrong-to-right transitions come from the ${\sim}65$ items that receive more than two loops, confirming that the extra computation is responsible for the accuracy gain. The $3$ right-to-wrong cases are items where over-looping past the correct answer at depth two leads to a changed and incorrect response, but the oracle-distilled head keeps such regressions rare. The pattern holds across all three training seeds.

\begin{table}[t]
\centering
\small
\setlength{\tabcolsep}{6pt}
\begin{tabular*}{\columnwidth}{@{\extracolsep{\fill}}l c c c@{}}
\toprule
 & \multicolumn{2}{c}{\textbf{\boldmath Fixed $T{=}2$}} \\
\cmidrule(lr){2-3}
\textbf{Adaptive} & \textbf{Correct} & \textbf{Wrong} & \textbf{Total} \\
\midrule
Correct & 1310 & 56 & 1366 \\
Wrong & 3 & 1031 & 1034 \\
\midrule
Total & 1313 & 1087 & 2400 \\
\bottomrule
\end{tabular*}
\caption{Paired transition table of per-item correctness under adaptive halting ($\tau{=}0.50$, $T_{\min}{=}2$) versus fixed $T{=}2$ (three-training-seed means, rounded). The $56$ wrong-to-right flips and $3$ right-to-wrong flips yield a net $+53$. Exact McNemar $p < 0.001$ on every training seed.}
\label{tab:transition}
\end{table}

\paragraph{Halting-Head Calibration and an Oracle-Depth Bound.}
The counts here are three-training-seed means; because each seed deepens a different set, they are per-seed averages rather than one fixed set of items. The head deepens about $65$ items per training seed and about $56$ flip from wrong to right. Two distinct rates describe the deepened items. The oracle-label precision is about $85\%$: roughly $55$ of the $65$ are oracle-positive, their loss falling by more than the margin $\delta$. The answer-flip rate is about $86\%$: roughly $56$ of the $65$ turn a wrong greedy answer right. Both rates are best read against their population base rates. At depth two about $153$ of the $2400$ items ($6.4\%$) are oracle-positive, so the $85\%$ precision on the selected tail is roughly thirteen times the population rate, at a recall of about $36\%$ ($55$ of $153$). The flip rate has a tighter reference. Of the $1087$ items that fixed $T{=}2$ answers wrongly, only $151$ ($13.9\%$) are answered correctly at any depth in $1$--$16$, a count implied by the correctness ceiling below ($61.0\%$ is $1464$ of $2400$, against $1313$ at fixed $T{=}2$). Deepening a randomly chosen subset of the wrong items therefore flips at most $13.9\%$ of them, so the head's $86\%$ is at least six times what blind deepening reaches, and it recovers $56$ of the $151$ recoverable items while touching only $65$. This is consistent with the per-depth AUROC of $0.74$ in Table~\ref{tab:depth-cal}, which summarizes ranking quality averaged over the whole distribution and places no bound on precision in the extreme tail that a threshold firing on $2.7\%$ of items reads. The two rates need not coincide, since a supra-$\delta$ loss drop and a changed argmax are distinct criteria, but at this operating point both signals concentrate in the same tail. As an accuracy upper bound, a \emph{correctness} oracle that assigns each item the depth at which its greedy answer is correct (if any depth yields a correct answer) reaches $61.0\%$. This is a different criterion from the loss oracle of Eq.~\eqref{eq:oracle}, which marks whether a deeper loop lowers the teacher-forced loss rather than whether it flips the greedy answer. RecurTrace sits at $54.7 < 56.9 < 61.0$, above the best fixed depth and below this correctness ceiling.

The same head architecture supports all four trained-halting baselines. With the ACT objective \citep{graves2016act} the head forms a halting distribution and pays an expected-depth ponder cost of $0.01$. With the PonderNet objective \citep{banino2021pondernet} it pays a Kullback-Leibler penalty of weight $0.01$ toward a geometric prior with parameter $0.2$. Both objectives use the absolute distance to the oracle target depth as a per-depth surrogate quality cost, so they see the same labels as the oracle distillation. On the looped backbone both collapse to a single loop, as Table~\ref{tab:main} reports, because the per-example loss is already low after one loop and the compute penalty dominates the small fraction of inputs that improve with depth. To rule out that this collapse is an artifact of a single untuned penalty, we also swept the ACT ponder cost and the PonderNet KL weight over $\{0.001, 0.01, 0.1\}$ each, training only the halting head, and both stay collapsed at a mean depth of $1.0$ loops across all six settings. We read this collapse as a property of attaching penalty-based halting post hoc to a frozen looped backbone, where the head is fitted after the representation it reads is fixed and cannot shape the network so that deeper loops pay off, rather than as a refutation of ACT or PonderNet in the end-to-end settings they were designed for.

\paragraph{LoopUS-Conf and TaH-Mismatch baselines.}
Two additional baselines use the same halting head but replace our oracle signal. LoopUS-Conf follows the LoopUS confidence recipe \citep{park2026loopus}: the head is trained with a binary cross-entropy confidence loss plus a monotonicity regularizer that encourages the continue probability to decrease with depth, without any oracle labels. The confidence threshold is selected on held-out data as for RecurTrace. TaH-Mismatch adapts the oracle signal of Think-at-Hard \citep{fu2026tah} to the sequence level: the ``continue'' label at depth $t$ is $y_t=\mathbb{1}[\hat{a}_t \neq a]$, where $\hat{a}_t$ is the greedy answer decoded at depth $t$ and $a$ is the reference answer, and the head is trained with binary cross-entropy against this token-mismatch target. Both baselines use the same head capacity, optimizer, and threshold-selection protocol as RecurTrace, isolating the effect of the supervision signal.

\section{Benchmark Breakdown and Diagnostics}
\label{app:bench}
Table~\ref{tab:app-bench} lists the benchmark groups and diagnostic groups with item counts evaluated after bucket-stratified subsampling. The train-covered suite holds the fourteen tasks whose training splits enter the mixture, and the classic suite holds eight standard multiple-choice benchmarks. The held-out group, including the BIG-Bench Hard subsets, is diagnostic evidence only, not used for the main breadth claim. We evaluate with bucket-stratified sampling, reporting generation accuracy and teacher-forced NLL at loops in $\{1,2,4,6,8\}$.

\begin{table*}[t]
\centering
\small
\setlength{\tabcolsep}{6pt}
\begin{tabular}{l c p{0.62\textwidth} c}
\toprule
\textbf{Suite} & \textbf{\#} & \textbf{Benchmarks} & \textbf{Eval items} \\
\midrule
Train-covered & 14 & arithmetic, p-hop, symbolic, AQUA-RAT, bAbI, CLUTRR, GSM8K, HotpotQA, MATH, MathQA, MuSiQue, ProofWriter, StrategyQA, 2WikiMultiHopQA & 45{,}722 \\
\midrule
Diagnostic held-out & 8 & BBH date understanding, BBH logical deduction (three and five objects), BBH multistep arithmetic, BBH tracking shuffled objects, LAMBADA \citep{paperno2016lambada}, MultiArith \citep{roy2015multiarith}, SVAMP \citep{patel2021svamp} & 2{,}730 \\
\midrule
Classic & 8 & ARC-Easy, ARC-Challenge, OpenBookQA \citep{mihaylov2018openbookqa}, BoolQ \citep{clark2019boolq}, PIQA \citep{bisk2020piqa}, HellaSwag, WinoGrande \citep{sakaguchi2020winogrande}, MMLU & 4{,}785 \\
\bottomrule
\end{tabular}
\caption{Benchmarks and diagnostics; ``Eval items'' are the subsampled counts actually evaluated, not full test-set sizes. The BIG-Bench Hard held-out group is retained only as a diagnostic and is excluded from the main breadth claim reported in the body.}
\label{tab:app-bench}
\end{table*}

Table~\ref{tab:app-suite} reports per-suite changes against the same-budget baseline at the two scales with full per-suite generation, $0.6$B and $1.7$B. At $1.7$B the suite-level numbers match the breadth table in the body. Negative log-likelihood falls and accuracy rises on train-covered tasks and on the classic suite ($+1.9$ points). At $0.6$B the likelihood gains are the largest across scales because the weaker base has more headroom, with both suites showing meaningful NLL reductions. The overall same-budget accuracy gains at $4$B ($+2.2$ points) and $8$B ($+3.4$ points) appear in Table~\ref{tab:scaling}. The held-out and BIG-Bench Hard rows remain diagnostic and are not read as a main claim. For transparency the diagnostic scores are $+0.3$ pp on the BIG-Bench Hard subsets ($28.4\%$ to $28.7\%$) and $+0.4$ pp on the remaining held-out tasks ($34.8\%$ to $35.2\%$) at $1.7$B, weaker than the train-covered and classic results, consistent with treating them as out-of-scope diagnostics, not headline evidence.

\begin{table}[t]
\centering
\small
\setlength{\tabcolsep}{5pt}
\begin{tabular*}{\columnwidth}{@{\extracolsep{\fill}}l l c c@{}}
\toprule
\textbf{Scale} & \textbf{Suite} & \textbf{\boldmath NLL $\Delta$ $\downarrow$} & \textbf{\boldmath Acc $\Delta$ (pp) $\uparrow$} \\
\midrule
$0.6$B & Train-covered & $-0.11$ & $+0.3$ \\
$0.6$B & Classic & $-0.22$ & $+1.1$ \\
\midrule
$1.7$B & Train-covered & $-0.05$ & $+0.8$ \\
$1.7$B & Classic & $-0.13$ & $+1.9$ \\
\bottomrule
\end{tabular*}
\caption{Per-suite generation results at $0.6$B and $1.7$B against a same-budget baseline (three-training-seed means at both scales). Both NLL $\Delta$ (teacher-forced negative log-likelihood minus the no-loop baseline) and Acc $\Delta$ (generation accuracy, pp) are measured at a fixed two-loop depth, matching Table~\ref{tab:scaling}. The overall $4$B and $8$B same-budget gains appear in Table~\ref{tab:scaling}.}
\label{tab:app-suite}
\end{table}

\section{Difficulty-Adaptation Details}
\label{app:difficulty}
We split each seed's MathQA test items by correctness at a single loop. Easy items are already solved at loop depth one, and hard items, including items the model never answers correctly, need a deeper unroll (the easy/hard split is at loop depth one, where easy items are those answered correctly at $T{=}1$ and hard items are those answered incorrectly). This split is independent of the fixed two-loop baseline used for the paired transition counts, so the ${\sim}50$ hard items among the ${\sim}65$ deepened and the $56$ wrong-to-right flips against fixed two loops are counted against different references and do not conflict. The body uses two operating points from the same model and seeds, distinguished by the stopping threshold and the floor. Table~\ref{tab:reconcile} reconciles them.

\begin{table}[t]
\centering
\small
\setlength{\tabcolsep}{3pt}
\begin{tabular*}{\columnwidth}{@{\extracolsep{\fill}}l c c c c c@{}}
\toprule
\textbf{Setting} & \textbf{\boldmath$\tau$} & \textbf{\boldmath$T_{\min}$} & \textbf{Loops} & \textbf{Acc} & \textbf{Gap} \\
\midrule
Main (Tab.~\ref{tab:main}) & 0.50 & 2 & 2.04 & 56.9 & $+0.06$ \\
Difficulty (Fig.~\ref{fig:difficulty}) & 0.10 & none & 2.80 & 53.6 & $+0.59$ \\
\bottomrule
\end{tabular*}
\caption{Operating-point reconciliation. Both rows use the same model; the main row is the three-training-seed mean (Table~\ref{tab:main}) and the difficulty row the median training seed (Table~\ref{tab:app-thr}). The main point ($\tau{=}0.50$, $T_{\min}{=}2$) is the operating point selected on held-out NLL, and the difficulty point ($\tau{=}0.10$, no floor) maximizes the held-out gain and exposes the hard-minus-easy (H$-$E) gap more clearly. The difficulty point spends more compute ($2.80$ loops) yet reaches only $53.6\%$ ($1287$ of $2400$), below the best fixed depth, because without a floor it both stops some items too early and over-computes others, lowering accuracy despite the higher mean depth. It is an illustration of the adaptation pattern, not a competing accuracy number.}
\label{tab:reconcile}
\end{table}

\paragraph{Main operating point ($\tau{=}0.50$, $T_{\min}{=}2$).}
At the main operating point, the conservative threshold stops $97.3\%$ of items at the two-loop floor (Table~\ref{tab:depth-hist}). The hard-minus-easy loop gap is $+0.06$ loops, smaller than at $\tau{=}0.10$ but positive on all eight seeds ($+0.02$ to $+0.11$). Of the ${\sim}65$ items per training seed that receive more than two loops, ${\sim}50$ ($77\%$) are hard by the loop-depth-one split, so the halting head routes compute to items that need it. The paired transition breakdown (Table~\ref{tab:transition}) shows $56$ of these items flip from wrong to right and only $3$ from right to wrong, yielding the $+53$ net correct-answer gain. At this threshold the continue-probability distributions for oracle-positive and oracle-negative items remain separated on held-out data (Appendix~\ref{app:halting}), so a small mean depth gap can still select a high-precision tail.

\paragraph{Difficulty operating point ($\tau{=}0.10$).}
The difficulty figure reads the halting head at the deeper threshold ($\tau{=}0.10$, mean $2.80$ loops) that maximizes the held-out gain. Table~\ref{tab:app-seed} gives the per-seed mean depth on easy and hard items at this threshold. The halting head spends more loops on hard items than easy items on every one of the eight seeds, the gap averages $0.59$ loops, and the smallest per-seed gap is still positive.

\begin{table}[t]
\centering
\small
\setlength{\tabcolsep}{8pt}
\begin{tabular*}{\columnwidth}{@{\extracolsep{\fill}}l c c c@{}}
\toprule
\textbf{Seed} & \textbf{Easy loops} & \textbf{Hard loops} & \textbf{Gap} \\
\midrule
1 & 2.00 & 2.01 & $+0.01$ \\
2 & 2.77 & 4.12 & $+1.35$ \\
3 & 3.16 & 4.31 & $+1.15$ \\
4 & 1.97 & 2.12 & $+0.15$ \\
5 & 2.02 & 2.54 & $+0.51$ \\
6 & 2.42 & 3.02 & $+0.59$ \\
7 & 2.04 & 2.26 & $+0.22$ \\
8 & 3.65 & 4.40 & $+0.75$ \\
\midrule
Mean & 2.50 & 3.10 & $+0.59$ \\
\bottomrule
\end{tabular*}
\caption{Mean loop depth on easy and hard MathQA items per evaluation seed at the difficulty operating point ($\tau{=}0.10$), shown for the median training seed. Hard items receive more loops than easy ones on all eight evaluation seeds, and the mean gap is $+0.59$ loops. Gaps are computed from unrounded values. The pattern holds across all three training seeds.}
\label{tab:app-seed}
\end{table}

\begin{figure}[t]
\centering
\includegraphics[width=\columnwidth]{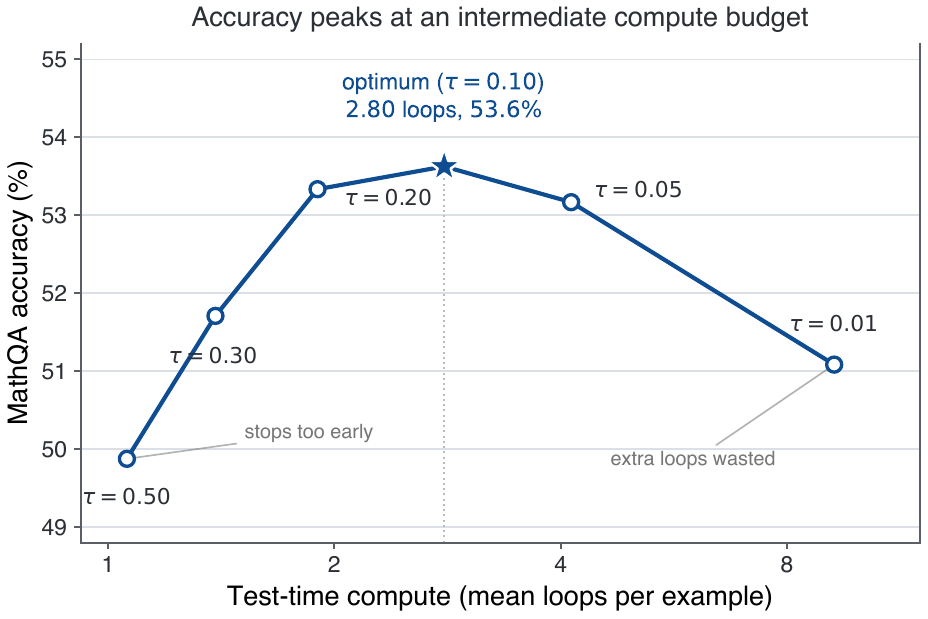}
\caption{Sweeping the shared stopping threshold $\tau$ without the $T_{\min}$ floor traces a compute-accuracy curve on MathQA. As $\tau$ falls the halting head keeps looping, so the mean loop count grows, and accuracy rises to an optimum near $\tau{=}0.10$ ($2.80$ loops, $53.6\%$), then declines as extra loops are wasted. The six points are the thresholds in Table~\ref{tab:app-thr}, accuracy recovered from the $\Delta$ correct column over the $2400$ MathQA items.}
\label{fig:threshold}
\end{figure}

Table~\ref{tab:app-thr} sweeps the shared threshold without the $T_{\min}{=}2$ floor and shows the trade it controls, which Figure~\ref{fig:threshold} plots as a compute-accuracy curve. A lower threshold keeps the halting head looping, which raises the mean depth, lowers the loop-count reduction, and widens the hard-easy gap. A higher threshold stops sooner. Across the safe range every threshold keeps a nonnegative held-out gain on all eight seeds, and $\tau{=}0.10$ is the value that maximizes the summed gain while keeping the hard-easy gap positive on every seed. At $\tau{=}0.50$ without a floor the mean depth drops to $1.06$ because the head is confident enough to stop most items after a single loop. Adding the $T_{\min}{=}2$ floor as in the main operating point forces every item to at least two loops and raises the mean to $2.04$, producing the accuracy reported in Table~\ref{tab:main}.

\begin{table}[t]
\centering
\small
\setlength{\tabcolsep}{6pt}
\begin{tabular*}{\columnwidth}{@{\extracolsep{\fill}}c c c c c@{}}
\toprule
\textbf{\boldmath$\tau$} & \textbf{Mean loop} & \textbf{Loop cut (\%)} & \textbf{\boldmath Hard$-$easy} & \textbf{\boldmath$\Delta$ correct} \\
\midrule
0.01 & 9.24 & 42.3 & $+1.82$ & $+85$ \\
0.05 & 4.13 & 74.2 & $+1.13$ & $+135$ \\
\textbf{0.10} & \textbf{2.80} & \textbf{82.5} & $\mathbf{+0.59}$ & $\mathbf{+146}$ \\
0.20 & 1.90 & 88.1 & $+0.29$ & $+139$ \\
0.30 & 1.39 & 91.3 & $+0.18$ & $+100$ \\
0.50 & 1.06 & 93.4 & $+0.03$ & $+56$ \\
\bottomrule
\end{tabular*}
\caption{Shared-threshold sweep on MathQA (eight evaluation seeds, median training seed). Loop cut is the mean loop-count reduction relative to a sixteen-loop budget ($1{-}\bar T/16$), not total compute. Hard$-$easy is the mean hard-minus-easy loop gap, and $\Delta$ correct is the summed gain in correct answers over that budget. The selected operating point ($\tau{=}0.10$) is in bold.}
\label{tab:app-thr}
\end{table}

\section{Additional Ablations and Discussion}
\label{app:discussion}
\paragraph{Memory Against Plain Looping.}
The consolidated ablation in Table~\ref{tab:memory} isolates Loop Memory Attention by removing it and looping the block as plain re-iteration. On the train-covered suite the teacher-forced NLL change against the same-budget baseline at the best-likelihood loop is $-0.06$ nats at $0.6$B and $+0.07$ nats at $1.7$B for plain looping, against $-0.15$ and $-0.03$ nats once the memory is added. At $4$B and $8$B the pattern strengthens: plain looping drifts to $+0.05$ and $+0.06$~nats while memory yields $-0.03$ and $-0.02$~nats. Plain looping on a stronger base drifts the representation without adding information, and memory turns the extra passes into an NLL reduction. The held-out and classic suites move the same direction at both scales, so the effect is not specific to training tasks. The generation-accuracy counterpart at $1.7$B, where Loop Memory Attention improves over the memoryless Encode-Think-Decode-style baseline by $2.3$ points on MathQA and $2.4$ on the classic suite, is in the same table. These ablation runs are trained separately from the breadth models, so their full-model NLL changes on the train-covered suite differ from the breadth measurements (Tables~\ref{tab:breadth} and~\ref{tab:app-suite}) by $0.02$ nats at $1.7$B and $0.04$ nats at $0.6$B.

\paragraph{Matched Same-Start Memory Control at $8$B.}
To test loop memory at the $8$B scale under a fixed budget, we start from a trained plain-looping checkpoint and continue training in two matched ways. One adds the loop-memory modules, a single-window memory with the gate and input injection trained without auxiliary losses, and the other continues plain looping for the same number of steps. The memory continuation reaches lower teacher-forced NLL than the matched plain continuation on both seeds, by $0.040$ and $0.028$ nats at the best loop and by $0.016$ and $0.050$ nats at four loops. Because the comparison shares the same starting checkpoint and the same extra budget, it isolates the memory from the additional training steps. We report it as a controlled single-configuration result rather than a broad necessity claim.

\paragraph{Task Type Decides Plain Against Memory.}
A controlled study on the synthetic extrapolation buckets shows the picture is not one-sided. On pure pointer-chasing chains, where the task is a clean fixed-point iteration, plain looping extrapolates a little better than the memory variant at deep unrolls, which trails only slightly rather than collapsing. On the stateful tasks that must carry intermediate values, the symbolic and arithmetic chains, the memory variant is the stronger of the two. This is the behavior the token-wise gate is designed to reconcile. It can close the memory on stateless patterns and open it on stateful ones, which is why a single setting holds across the breadth suites without per-task tuning. The per-bucket accuracies behind this split appear in the bottom block of Table~\ref{tab:memory} in the main text.

\paragraph{Parameter-Matched Non-Loop Control.}
A looped model adds about $2.2$ percent trainable parameters over the frozen base, so one alternative explanation for its gains is the extra capacity rather than the recurrence. To separate the two at $1.7$B we train a non-looped control that adds the same budget of trainable parameters, a low-rank adapter of matched size on the same block, but applies it only once with no looping, and we compare it against RecurTrace under identical data, steps, and evaluation. The matched-capacity control reaches $45.4\%$ generation accuracy against the $44.8\%$ same-budget baseline, so it captures about half of RecurTrace's $+1.2$-point gain, and adding recurrence and memory supplies the other half to reach $46.0\%$. The likelihood picture is more one-sided, since the capacity control lowers the overall $22$-task teacher-forced NLL by only $0.01$ nats against that baseline while RecurTrace lowers it by $0.08$. Repeated depth with memory, not added capacity alone, carries the effect.

\paragraph{Per-Scale Summary.}
Table~\ref{tab:scaling} collects the same-budget generation-accuracy and teacher-forced NLL changes across scales. Accuracy gains grow with scale, from $+0.6$ to $+3.4$ points, while the teacher-forced NLL reduction shrinks over the same range, from $0.15$ to $0.01$ nats. We read the two trends as consistent rather than conflicting, because the metrics average over different populations. Teacher-forced NLL is a mean over every answer token, and most of those tokens a strong base already predicts confidently, so the headroom a ${\approx}2\%$ module can move shrinks toward zero as the base improves and the mean barely shifts. Generation accuracy records only whether the argmax changes, so it is decided by the minority of items whose top two candidates sit close together, where a logit change too small to register in the mean NLL is enough to flip the answer. At $0.6$B the base is weak, so likelihood has broad headroom while many of the items it improves remain far from correct; at $8$B the balance reverses. The paragraphs below expand on the $4$B and $8$B rows.

\begin{table}[t]
\centering
\small
\setlength{\tabcolsep}{8pt}
\begin{tabular*}{\columnwidth}{@{\extracolsep{\fill}}l c c c c@{}}
\toprule
\textbf{Scale} & \textbf{Baseline Acc} & \textbf{Loop Acc} & \textbf{\boldmath Acc $\Delta$ $\uparrow$} & \textbf{\boldmath NLL $\Delta$ $\downarrow$} \\
\midrule
$0.6$B & 35.2 & 35.8 & $+0.6$ & $-0.15$ \\
$1.7$B & 44.8 & 46.0 & $+1.2$ & $-0.08$ \\
$4$B & 58.6 & 60.8 & $+2.2$ & $-0.02$ \\
$8$B & 66.6 & 70.0 & $+3.4$ & $-0.01$ \\
\bottomrule
\end{tabular*}
\caption{Overall $22$-task (unweighted task mean) generation accuracy (\%) and teacher-forced NLL change (nats) over the same-budget fine-tuned baseline at each scale, at a fixed two-loop depth, shown to one decimal. Every row averages $3$ training seeds (std $0.22$, $0.29$, $0.33$, $0.37$ pp from $0.6$B to $8$B). Per-suite and per-task breakdowns appear in Tables~\ref{tab:app-suite},~\ref{tab:8b-detail}, and~\ref{tab:8b-tasks}.}
\label{tab:scaling}
\end{table}

\paragraph{The $4$B Comparison.}
At $4$B, RecurTrace improves generation accuracy by $2.2$ points over a fine-tuned baseline trained under the same budget, and it lowers teacher-forced NLL by about $0.02$ nats. The gain is consistent with the smaller scales and is measured against the same-budget baseline rather than a per-task loop oracle. The $300$-step module follow-up is shorter than the $0.6$B and $1.7$B schedules, so we treat this row as an initial scale check rather than a fully matched training-length replication. The halting head is distilled at $4$B as at the other scales (Stage~2, Table~\ref{tab:budget}); we report this scaling row at a fixed two-loop depth and leave a full adaptive-halting \emph{evaluation} at $4$B to future work.

\paragraph{The $8$B Comparison.}
At $8$B, RecurTrace improves generation accuracy by $3.4$ points over a same-budget fine-tuned Qwen3-8B baseline. Table~\ref{tab:8b-detail} reports the full per-suite breakdown with absolute accuracies across three seeds. The baseline fine-tunes all attention and feed-forward parameters of Qwen3-8B-Base on the same reasoning mixture for $800$ optimizer steps using eight H800 GPUs, bfloat16 precision, a global batch of $128$, a peak learning rate of $1\mathrm{e}{-}5$ with a cosine schedule, and a maximum length of $1024$ tokens. RecurTrace freezes the base and trains only the loop, memory, and injection modules ($\approx$0.12B, at most $2\%$) for the same $800$ steps with the same optimizer and schedule (Stage~1); the halting head is distilled separately afterwards (Stage~2, Appendix~\ref{app:halting}). Both runs therefore consume about $105$M training tokens ($800$ steps $\times$ $128$ sequences $\times$ $1024$ tokens), each on the same hardware and precision, with the training budget accounted for in Table~\ref{tab:budget}: because the baseline updates all $\approx$8.2B parameters while RecurTrace updates only the added modules, any budget mismatch favors the baseline, so the gain is not an artifact of extra training compute. The comparison replaces an earlier one against the untuned Qwen3-8B-Base, which overstated the gap by also measuring the distance from an untrained checkpoint to any supervised model. RecurTrace at $8$B is evaluated at a fixed two-loop depth; the halting head is not invoked in this scaling comparison, so the reported gain reflects the loop-memory mechanism. Teacher-forced NLL drops by $0.01$ nats at $8$B, a small but consistent improvement.

\begin{table}[t]
\centering
\small
\setlength{\tabcolsep}{4pt}
\begin{tabular*}{\columnwidth}{@{\extracolsep{\fill}}l c c c@{}}
\toprule
\textbf{Suite} & \textbf{Baseline} & \textbf{RecurTrace} & \textbf{\boldmath$\Delta$ (pp) $\uparrow$} \\
\midrule
Train-covered (14 tasks) & 62.3 & 66.9 & $+4.6$ \\
Classic (8 tasks) & 74.0 & 75.4 & $+1.4$ \\
\midrule
Overall (22 tasks) & 66.6 & 70.0 & $+3.4$ \\
\bottomrule
\end{tabular*}
\caption{$8$B per-suite generation accuracy (\%) over three seeds. The baseline is a same-budget fine-tuned Qwen3-8B. Overall is the unweighted mean over all $22$ tasks, i.e., the task-count-weighted mean of the two suite rows. Per-seed overall gains are $+3.02$, $+3.69$, $+3.61$ (mean $3.4$, std $0.37$ pp). Teacher-forced NLL drops by $0.01$ nats.}
\label{tab:8b-detail}
\end{table}

Table~\ref{tab:8b-tasks} gives the per-task breakdown for selected benchmarks at $8$B. Among the displayed train-covered tasks, gains range from $+3.4$ to $+5.3$ pp, and across all $14$ train-covered tasks they range from $+1.7$ to $+6.1$ pp, while classic commonsense benchmarks are more mixed ($-0.6$ to $+3.1$ pp), and two of them (BoolQ, WinoGrande) show small negative changes within seed noise. This pattern is consistent with a mechanism that adds latent reasoning depth rather than general capacity. Per-task three-seed standard deviations range from $0.4$ to $1.9$ pp, and the overall per-seed standard deviation is $0.37$ pp.

\begin{table*}[t]
\centering
\small
\setlength{\tabcolsep}{5pt}
\begin{tabular*}{\textwidth}{@{\extracolsep{\fill}}l c c c | l c c c@{}}
\toprule
\multicolumn{4}{c|}{\textbf{Train-covered tasks}} & \multicolumn{4}{c}{\textbf{Classic tasks}} \\
\textbf{Task} & \textbf{Base} & \textbf{Loop} & \textbf{\boldmath$\Delta$ $\uparrow$} & \textbf{Task} & \textbf{Base} & \textbf{Loop} & \textbf{\boldmath$\Delta$ $\uparrow$} \\
\midrule
GSM8K & 44.8 & 50.1 & $+5.3$ & ARC-Easy & 82.3 & 84.0 & $+1.7$ \\
MATH & 31.6 & 36.4 & $+4.8$ & ARC-Challenge & 69.5 & 72.6 & $+3.1$ \\
MathQA & 67.5 & 72.6 & $+5.1$ & OpenBookQA & 70.1 & 71.9 & $+1.8$ \\
AQUA-RAT & 51.8 & 55.9 & $+4.1$ & BoolQ & 79.8 & 79.4 & $-0.4$ \\
HotpotQA & 54.6 & 59.8 & $+5.2$ & PIQA & 81.6 & 82.1 & $+0.5$ \\
bAbI & 89.2 & 92.6 & $+3.4$ & HellaSwag & 74.8 & 76.9 & $+2.1$ \\
ProofWriter & 79.4 & 83.7 & $+4.3$ & WinoGrande & 71.6 & 71.0 & $-0.6$ \\
 & & & & MMLU & 62.4 & 65.4 & $+3.0$ \\
\bottomrule
\end{tabular*}
\caption{Per-task generation accuracy (\%) at $8$B, three-seed mean. Base is the same-budget fine-tuned Qwen3-8B baseline, and Loop is RecurTrace. Multi-step reasoning tasks gain the most, while two classic benchmarks show small negative changes within seed noise. The remaining seven train-covered tasks (arithmetic, p-hop, symbolic, CLUTRR, MuSiQue, StrategyQA, 2WikiMultiHopQA) range from $+1.7$ to $+6.1$ pp. Per-task three-seed standard deviations range from $0.4$ to $1.9$ pp.}
\label{tab:8b-tasks}
\end{table*}

\paragraph{Reproducibility.}
The looped model lives in the local Qwen3 modeling code and loads on top of standard Qwen3 checkpoints, so the base weights load unchanged and only the loop parameters initialize fresh. All training uses eight H800 GPUs. The evaluation sweeps loops in $\{1,2,4,6,8\}$ with bucket-stratified sampling and reports both generation accuracy and teacher-forced negative log-likelihood. The MathQA adaptive comparison aggregates eight seeds of three hundred items each, for $2400$ test items, and selects every threshold on held-out data. The block-selection probe, the synthetic data generators, and the evaluation scripts are deterministic given their seeds and configurations.

\end{document}